\documentclass[sigconf]{acmart}
\DeclareMathOperator*{\argmax}{arg\,max}
\usepackage{multirow}
\usepackage{booktabs}
\usepackage{makecell}
\usepackage{array}
\usepackage{arydshln}
\usepackage{graphicx}  % 核心：插入图片
\usepackage{subcaption}% 核心：提供 subfloat 环境，解决子图排版
\usepackage{enumitem}
\usepackage{bm}
\usepackage[normalem]{ulem}
\useunder{\uline}{\ul}{}

\AtBeginDocument{%
  }

\copyrightyear{2026}
\acmYear{2026}
\setcopyright{cc}
\setcctype{by-nc-nd}
\acmConference[RecSys '26]{20th ACM Conference on Recommender Systems}{September 27-October 02, 2026}{Minneapolis, MN, USA}
\acmBooktitle{20th ACM Conference on Recommender Systems (RecSys '26), September 27-October 02, 2026, Minneapolis, MN, USA}
\acmDOI{10.1145/3773078.3831777}
\acmISBN{979-8-4007-2284-4/2026/09}

\usepackage{xspace}

\begin{document}


%\title{The Name of the Title Is Hope}
\title{Empowering Cross-Domain Sequential Recommendation with Hybrid Tokenization and Serial-Parallel Decoding}

\author{Yuxuan Hu}
%\thanks{\dagger \text{Co-first author}}
\affiliation{%
  \institution{City University of Hong Kong}
  \city{Hong Kong}
  \country{China}
}
\email{yuxuanhu7-c@my.cityu.edu.hk}

\author{Yuhao Wang}
%\thanks{\dagger \text{Co-first author}}
\affiliation{%
  \institution{City University of Hong Kong}
  \city{Hong Kong}
  \country{China}
}
\email{yhwang25-c@my.cityu.edu.hk}

\author{Tianbo Huang}
\affiliation{%
  \institution{ByteDance Inc.}
  \city{Beijing}
  \country{China}
}
\email{huangtianbo.1@bytedance.com}

\author{Chao Zhang} %\Letter}
\affiliation{%
  \institution{City University of Hong Kong}
  \city{Hong Kong}
  \country{China}
}
\email{czhang2328-c@my.cityu.edu.hk}

\author{Ziwei Liu} %\Letter}
\affiliation{%
  \institution{City University of Hong Kong}
  \city{Hong Kong}
  \country{China}
}
\email{lziwei2-c@my.cityu.edu.hk}

\author{Lihua Zhang} % \Letter
%\thanks{\Letter \text{Corresponding author}}
\affiliation{%
  \institution{ByteDance Inc.}
  \city{Beijing}
  \country{China}
}
\email{lizhiyu.0@bytedance.com}

\author{Xiangyu Zhao}
\authornote{Corresponding author.}
\affiliation{%
  \institution{City University of Hong Kong}
  \city{Hong Kong}
  \country{China}
}
\email{xianzhao@cityu.edu.hk}
\renewcommand{\shortauthors}{Yuxuan Hu et al.}

\begin{abstract}
%跨域序列推荐（CDSR）需要在多领域异质物品空间中建模用户兴趣的动态迁移。尽管生成式推荐将推荐统一为自回归序列生成，但现有方法往往在表示层面对跨域协同信息建模不足，并在推理阶段面临显著的解码延迟。为此，我们提出 GenCDSR，一个兼顾效果与效率的生成式跨域序列推荐框架。GenCDSR 通过 Share–Specific 跨域 Tokenization 学习领域感知的语义物品标识符：在标识符构建中联合引入共享与领域特定码本，从而同时捕获跨域语义共性与领域特性；在生成与推理方面，我们设计 Hybrid Serial–Parallel Decoding，在保证生成一致性的同时将跨域生成过程部分并行化，显著降低推理延迟。我们在三个公共数据集上的实验表明，GenCDSR 在推荐性能上稳定优于多种强基线，并在推理效率方面取得显著提升。

%Cross-domain sequential recommendation (CDSR) aims to model users’ dynamic interest transitions across multiple heterogeneous item spaces. 
Cross-domain sequential recommendation (CDSR) aims to model users’ dynamic interest transitions and sequential patterns across multiple domains. 
%to simultaneously improve their recommendation performance. 
%Although generative recommendation unifies the recommendation process across different domains as autoregressive sequence generation, existing approaches often fall short in capturing cross-domain collaborative signals at the representation level and incur substantial decoding latency during inference. 
Recently, generative recommendation (GR) has emerged, which first learns semantic identifiers (SIDs) using semantic information of items and models the recommendation task as autoregressive generation.
However, it faces two critical issues:
1) ignoring collaborative correlations across different domains in tokenization step and 2) adopting inefficient decoding strategies like beam search in generation step, which hinders GR's application in real-time services.
% To address these issues, we propose GenCDSR, an effective and efficient generative framework for CDSR. 
% Specifically, GenCDSR learns domain-aware SIDs through a cross-domain hybrid tokenization mechanism, which jointly incorporates domain-shared and domain-specific codebooks to capture both cross-domain commonalities and distinctions. 
% %For generation and inference, 
% Furthermore, we design a serial–parallel decoding strategy that partially parallelizes cross-domain generation while preserving generation consistency, thus significantly reducing inference latency. 
To address these limitations, we propose GenCDSR, an effective and efficient generative framework for CDSR. Specifically, we design a cross-domain hybrid tokenization mechanism that employs a multi-tower architecture to jointly capture cross-domain commonalities and domain-specific distinctions through hierarchical shared-specific and fine-grained codebooks. Furthermore, we develop a cross-domain serial-parallel decoding strategy that leverages the hierarchical SID structure to partially parallelize generation, significantly reducing inference latency while preserving generation consistency.
%Experimental results on three public datasets validate that GenCDSR consistently outperforms SOTA baselines in recommendation accuracy and achieves substantial gains in inference efficiency.
Experimental results on three public datasets validate that GenCDSR achieves a 1.5\% improvement in accuracy and an 85.1\% reduction in inference latency on average compared to SOTA baselines.
The implementation code and datasets are available online: \url{https://github.com/Applied-Machine-Learning-Lab/RecSys2026_GenCDSR}.
  
\end{abstract}


%\ccsdesc[500]{Information systems}
%\ccsdesc[500]{Information systems~Recommender systems}

%\ccsdesc[500]{Information systems}


\maketitle

\section{Introduction}

Cross-domain sequential recommendation (CDSR)~\cite{liu2025llm,liu2025bridge} models users' interaction sequences across multiple commercial domains, such as product categories on e-commerce and online entertainment platforms~\cite{cao2022contrastive,xu2024rethinking,hu2026emotion,chen2024tgca}, to capture dynamic interest transitions across heterogeneous item spaces and improve recommendation performance. A representative line of research models mixed behavioral sequences. For example, TriCDR~\cite{ma2024triple} extracts shared interests through multi-granularity attention and cross-domain contrastive learning. Meanwhile, generative recommendation (GR) reformulates recommendation as autoregressive generation. Existing GR frameworks~\cite{rajput2023recommender,wang2024learnable} typically involve two steps: 1) \textbf{Tokenization}, where a quantization model such as RQ-VAE~\cite{lee2022autoregressive} maps item semantic embeddings into sequences of semantic identifiers (SIDs) to capture item correlations; and 2) \textbf{Generation}, where next-token prediction leverages language models to capture temporal patterns and generate the SID of the next item. For example, TIGER~\cite{rajput2023recommender} first learns SIDs through RQ-VAE tokenization and then trains a seq2seq model to generate the next-item SID.

%Despite these advances, standard GR approaches often treat different domains as isolated silos, largely overlooking the potential for cross-domain collaborative correlation. 
%To address this, recent efforts have begun to explore cross-domain semantic generative recommendation, aiming to establish shared semantic anchor spaces across different domains. through semantic association mechanisms.

% 跨域生成式推荐面临两方面的核心挑战
% 第一，只考虑了不同物品的语义关联，fail to capture the complex correlation between domains from collaborative（这里需要分析实验的证据）。
% 第二，生成式推荐的token by token解码部分（以beam search为代表的）推理时延过高（这里放实验章节Inference Efficiency的图片），hinder生成式模型上线。

% 尽管生成式范式为跨域序列推荐提供了统一建模的可能，但将现有生成式方法直接迁移到跨域场景仍面临两项关键挑战。第一，表示层面的跨域协同信息建模不足。现有跨域生成式方法往往侧重于物品语义关联或使用单一量化器进行跨域离散化，这会使跨域交互所蕴含的协同结构难以在语义 ID 中得到充分编码，从而导致离散表示的信息损失与泛化受限。我们在特征保真度实验中观察到这一点：如图~\ref{fig:fidelity} 所示，相比跨域共享单一 RQ-VAE 或“每域各自 RQ-VAE”的方案，我们的方法在三个跨域数据集上均取得更高的 Feature Fidelity，说明其生成的语义 ID 能更充分保留原始语义特征，为后续跨域建模提供更可靠的表示基础。第二，推理阶段的逐 token 串行解码效率瓶颈。生成式推荐通常依赖 token-by-token 的自回归解码（例如 Beam Search），推理时延随解码步骤与候选规模迅速攀升，成为模型上线部署的主要障碍；这一现象在我们的效率对比中同样得到验证：如表~\ref{tab:decoding} 所示，Beam Search 的 LT 显著高于其他解码策略，而更高效的并行类方法（如 MTP）又往往牺牲推荐效果。上述证据表明，跨域生成式推荐不仅需要在表示学习阶段显式兼顾共享语义与领域差异并注入跨域协同信息，还需要在推理阶段设计与两阶段语义 ID 结构匹配的高效解码机制，才能同时实现推荐效果与部署效率的可用平衡。
Although GR possesses great potential to improve CDSR, we find that
directly applying GR on CDSR setting faces two key challenges. First, in tokenization step cross-domain collaborative correlations tend to be insufficiently modeled~\cite{hu2026ids,jin2025generative}. %at the representation level. 
This is because most existing CDSR approaches 
%~\cite{hu2026ids,jin2025generative}
either train a separate quantization model for each domain~\cite{rajput2023recommender} or adopt a unified shared quantization model indiscriminately for all domains~\cite{jin2025generative},
%emphasize semantic associations among items or rely on a single quantizer shared across domains, 
making it difficult to explicitly encode the underlying collaborative correlations from cross-domain interactions into SIDs.
This limitation would further lead to information loss in discretized representations, thereby restricting the model’s generalization ability.
We will further analyze this phenomenon in the feature fidelity analysis in Section~\ref{sec:ffa}.
%We observe this issue in our feature-fidelity study: as shown in Figure~\ref{fig:feature fidelity}, our method consistently achieves higher Feature Fidelity than both a single shared RQ-VAE and domain-wise independent RQ-VAE across three cross-domain datasets, indicating that the resulting SIDs preserve the original semantic features more effectively and provide a more reliable representational basis for cross-domain modeling. 
Second, the generation step either suffers from efficiency bottleneck of token-by-token serial decoding~\cite{rajput2023recommender,zheng2024adapting} or degraded recommendation quality under fully parallel decoding~\cite{freitag2017beam,wang2025nezha}.
Specifically, the inference latency grows rapidly with decoding steps and candidate expansion in the widely adopted serial decoding strategies like Beam Search~\cite{freitag2017beam}, 
%Generative recommendation typically relies on autoregressive decoding 
%(e.g., Beam Search), 
posing a major obstacle to practical deployment on real-time services. 
Besides, common parallel decoding strategies like Multi-token Prediction (MTP)~\cite{gloeckle2024better} may predict multiple positions independently and fail to leverage semantic dependencies from preceding tokens, resulting in lower per-position accuracy and degraded recommendation quality.
We will provide concrete evidence in Section~\ref{sec:efficiency}.
%This trend is also reflected in our efficiency comparison: as shown in Table~\ref{tab:Component analysis}, Beam Search exhibits substantially higher latency (LT) than other decoding strategies, while more efficient parallel-style methods (e.g., MTP) often compromise recommendation quality. Taken together, these observations suggest that effective generative cross-domain recommendation requires not only representation learning that explicitly accounts for shared semantics and domain-specific variations while injecting cross-domain collaboration, but also an efficient decoding mechanism aligned with the two-stage SID structure to achieve a practical balance between effectiveness and deployment efficiency.

% 因此，为了应对上述挑战，本文提出了一种高效的生成式跨域序列推荐框架 GenCDSR，从跨域协同信息提取与推理效率两个方面对生成式跨域推荐进行系统性建模。从表示层面看，GenCDSR 通过学习域感知的语义物品标识符（domain-aware semantic IDs），在标识符构建过程中同时引入共享代码本与域特定代码本，从而在统一的生成式框架下有效刻画跨域共性语义与域内差异特征。这种共享–特定相结合的设计为跨域语义对齐提供了结构化基础，有助于提升生成式序列建模的表达能力。
% 从生成与推理层面看，本文进一步针对域感知语义标识符设计了一种混合串并行解码策略，使得不同领域的生成过程能够利用并行-串行结合的方式执行。相比传统逐令牌的纯串行解码方式，该策略在保持生成推荐效果的同时显著降低了推理时延，从而有效提升了模型的实用性。我们在两个真实跨域数据集上进行了大量实验，结果表明，GenCDSR 在推荐性能方面优于现有单域与跨域基线方法，并在推理效率上取得了显著提升。
% 本文的主要贡献总结如下：

% 提出了一种新的生成式跨域序列推荐框架 GenCDSR，在统一的生成式建模范式下同时刻画跨域语义共性与域内特征差异。
% 提出了一种面向域感知语义标识符的混合串并行解码策略，显著降低了生成式推荐的推理时延。
% 在多个真实数据集上的实验结果验证了 GenCDSR 在推荐效果与推理效率方面的优越性。

Therefore, to address these challenges, we propose GenCDSR, an effective and efficient generative cross-domain sequential recommendation framework. 
%that systematically models generative cross-domain recommendation from the perspectives of cross-domain collaborative information extraction and inference efficiency. 
%From the representation perspective, 
On the one hand, in the tokenization step, GenCDSR learns domain-aware SIDs, where both shared and domain-specific codebooks are jointly incorporated during SIDs training. This design enables the unified modeling of cross-domain semantic commonalities and domain-specific distinctions within a generative framework, enhancing the expressive capacity of generative sequence modeling.
On the other hand, in the generation step, we further design a cross-domain serial-parallel decoding strategy, tailored to domain-aware SIDs, allowing the generation process across different domains to be partially parallelized in a serial-parallel manner. Compared with conventional token-by-token serial decoding and parallel decoding, the proposed strategy significantly can reduce inference latency while maintaining recommendation quality, thereby improving the practical applicability of GR models. 
% Extensive experiments on three cross-domain datasets demonstrate that GenCDSR consistently outperforms existing single-domain and cross-domain baselines in terms of recommendation performance, while achieving substantial improvements in inference efficiency.

The main contributions of this work are summarized as follows:

\begin{itemize}[leftmargin=*]
    \item We propose GenCDSR, an effective and efficient generative cross-domain sequential recommendation framework that models both cross-domain commonalities and domain-specific distinctions under a unified generative paradigm.
    \item We propose cross-domain hybrid tokenization mechanism featuring share-specific multi-tower RQ-VAE and cross-domain serial-parallel decoding strategy featuring two-pass state-carry serial-parallel SID decoding. %for domain-aware SIDs  %significantly reduces inference latency for generative recommendation.
    \item Experiments on three public cross-domain datasets demonstrate that GenCDSR outperforms SOTA baselines by 1.5\% in accuracy while achieving an 85.1\% reduction in inference latency.
    %in terms of both recommendation effectiveness and inference efficiency.
\end{itemize}

\section{Preliminary}
\subsection{Problem Definition}
In this paper, we focus on the CDSR problem and use a dual-domain setting as an example.  
Let $\mathcal{U} = \{u_1, u_2, \ldots, u_{|\mathcal{U}|}\}$ denote the set of users, and $|\mathcal{U}|$ is the total number of users. 
We denote the two domains as $A$ and $B$, with corresponding item sets $\mathcal{A}$ and $\mathcal{B}$, respectively.
For a user $u \in \mathcal{U}$, their historical interactions in each domain are represented as chronologically ordered sequences:
\begin{equation}
\label{eq:problem_user}
S_u^{A} = (a_1, a_2, \ldots, a_{n_A}), \quad
S_u^{B} = (b_1, b_2, \ldots, b_{n_B}),
\end{equation}
where $a_i \in \mathcal{A}$ and $b_j \in \mathcal{B}$ denote interacted items in domains $A$ and $B$, respectively, and $n_A$ and $n_B$ denote the corresponding sequence lengths in the two domains.

To characterize users’ cross-domain behaviors, interactions from both domains are merged according to their temporal order into a unified cross-domain behavior stream:
\begin{equation}
\label{eq:problem_behavior}
\bar{S}_u = (v_1, v_2, \ldots, v_{n}), \quad v_t \in \mathcal{A} \cup \mathcal{B},
\end{equation}
where $n = n_A + n_B$ denotes the total number of historical interactions of user $u$ across the two domains.

The objective of CDSR is to predict the next item that a user $u$ is most likely to interact with in each domain, given the cross-domain behavior stream as well as the corresponding domain-specific interaction histories. Formally, the task can be formulated as the following two prediction objectives:
\begin{align}
\label{eq:problem_objective}
\hat{a} &= \arg\max_{a \in \mathcal{A}} P(v_{n+1}=a \mid \bar{S}_u, S_u^{A}, S_u^{B}),\\
\hat{b} &= \arg\max_{b \in \mathcal{B}} P(v_{n+1}=b \mid \bar{S}_u, S_u^{A}, S_u^{B}).
\end{align}

% Residual Quantized Variational Autoencoder（RQ‑VAE）是一种用于将连续向量表示映射为离散语义标识符的分层量化方法。给定物品的连续语义表示，RQ‑VAE 首先通过编码器将其映射至潜在空间：
% 随后，RQ‑VAE 采用多级残差量化机制逐层离散化潜在表示。在第 l层量化中，从码本中选取与当前残差向量最近的代码：
% 并更新残差表示：
% 经过 L层量化后，原始连续表示被编码为一个离散语义标识符序列。该离散表示可作为生成式模型中的输入令牌，并通过解码器近似重构原始语义表示。RQ‑VAE 通常通过重构损失与量化损失的联合优化目标进行训练。
\subsection{RQ-VAE}

Residual Quantized Variational Autoencoder (RQ-VAE)~\cite{lee2022autoregressive} %a hierarchical discretization approach designed 
aims to map continuous representations into SIDs.
% Given a continuous semantic embedding $\mathbf{x} \in \mathbb{R}^D$, an encoder maps it into a latent representation:
% \begin{equation}
% \label{equ:RQVAE_enc}
% \mathbf{z}^{(0)} = \mathrm{Enc}(\mathbf{x}).
% \end{equation}
%RQ-VAE performs an $L$-level residual quantization procedure to discretize continuous item representations into semantic IDs (SIDs). 
Specifically, given an input embedding $\mathbf{x}$, an encoder first maps it into a latent semantic vector $\mathbf{z}$, which is then quantized with $L$ level-wise codebooks. At each level $l=1,\ldots,L$, the model selects a codeword from the codebook $\mathcal{C}^{l}=\{\mathbf{c}^{l}_1,\ldots,\mathbf{c}^{l}_K\}$, where $K$ denotes the codebook size, by minimizing the $\ell_2$ distance to the current residual $\mathbf{r}^{l-1}$. Formally, the SID $s^{l}$ is obtained as:
\begin{equation}
\label{equ:RQVAE_distance}
s^{l} = \arg\min_j \left\| \mathbf{r}^{l-1} - \mathbf{c}^{l}_j \right\|_2,
\end{equation}
and the residual is updated accordingly:
\begin{equation}
\label{equ:RQVAE_residual}
\mathbf{r}^{l} = \mathbf{r}^{l-1} - \mathbf{c}^{l}_{s^{l}}.
\end{equation}

After $L$ quantization levels, the original embedding is represented by a length-$L$ discrete SID sequence $\{s^{1}, \ldots, s^{L}\}$. The corresponding quantized latent embedding is computed by aggregating the selected codewords, i.e.,
$\hat{\mathbf{z}} = \sum_{l=1}^{L} \mathbf{c}^{l}_{s^{l}}$,
which is further decoded into $\hat{\mathbf{x}}$ to reconstruct the input embedding $\mathbf{x}$. RQ-VAE is trained by jointly minimizing a reconstruction loss $\mathcal{L}_{\mathrm{recon}}$ and a residual quantization loss $\mathcal{L}_{\mathrm{RQ}}$, formulated as:
\begin{align}
\label{equ:RQVAE_loss}
&\mathcal{L}
=
\mathcal{L}_{\mathrm{recon}} + \mathcal{L}_{\mathrm{RQ}},\\
&\mathcal{L}_{\mathrm{recon}} = \left\| \mathbf{x} - \hat{\mathbf{x}} \right\|_2^2,\\
&\mathcal{L}_{\mathrm{RQ}}
=
\sum_{l=1}^{L}
\left(
\left\| \mathrm{sg}(\mathbf{r}^{l-1}) - \mathbf{c}^{l}_{s^{l}} \right\|_2^2
+
\beta
\left\| \mathbf{r}^{l-1} - \mathrm{sg}(\mathbf{c}^{l}_{s^{l}}) \right\|_2^2
\right),
\end{align}
where $\mathrm{sg}(\cdot)$ denotes the stop-gradient operator and $\beta$ controls the balance between codebook learning and encoder updates.

\section{Method}
% We first provide an overview of our GenCDSR framework, introduce the cross-domain hybrid tokenization and serial-parallel decoding modules, and present the training and inference procedures.

\subsection{Overview}

%为了解决现有生成式多场景推荐方法在表示建模和推理阶段面临的两项关键挑战——跨场景协同信息建模不足以及解码效率随场景数量增加而显著下降——本文提出了一种高效的生成式多场景推荐框架 GenCDSR。该框架主要由两个核心模块构成，如图【】所示。
% Existing generative cross-domain recommendation methods~\cite{hu2026ids,jin2025generative} suffer from insufficient modeling of cross-domain collaborative information at the representation level, primarily because item representations are either fully shared across domains or aligned only based on semantic similarity, without explicitly disentangling shared cross-domain semantics from domain-specific characteristics. As a result, collaborative signals across domains are weakly encoded in the learned representations and are instead left to be implicitly captured during subsequent sequence modeling, which limits effective cross-domain knowledge transfer. Moreover, these methods typically rely on token-by-token autoregressive decoding, leading to a rapid degradation of decoding efficiency.

% Existing generative cross-domain sequential recommendation methods~\cite{hu2026ids,jin2025generative} fail to explicitly disentangle shared and domain-specific factors, relying instead on fully shared or purely semantic representations. Furthermore, their token-by-token autoregressive decoding limits inference efficiency. To address these issues, we propose GenCDSR, an efficient generative framework comprising two core components: Cross-Domain Hybrid Tokenization and Cross-Domain Serial-Parallel Decoding. The overall framework is shown in Figure~\ref{fig:framework}.

\begin{figure*}
\setlength\abovecaptionskip{0.2\baselineskip}
\setlength\belowcaptionskip{0.2\baselineskip}
    \centering
    \includegraphics[width=\linewidth]{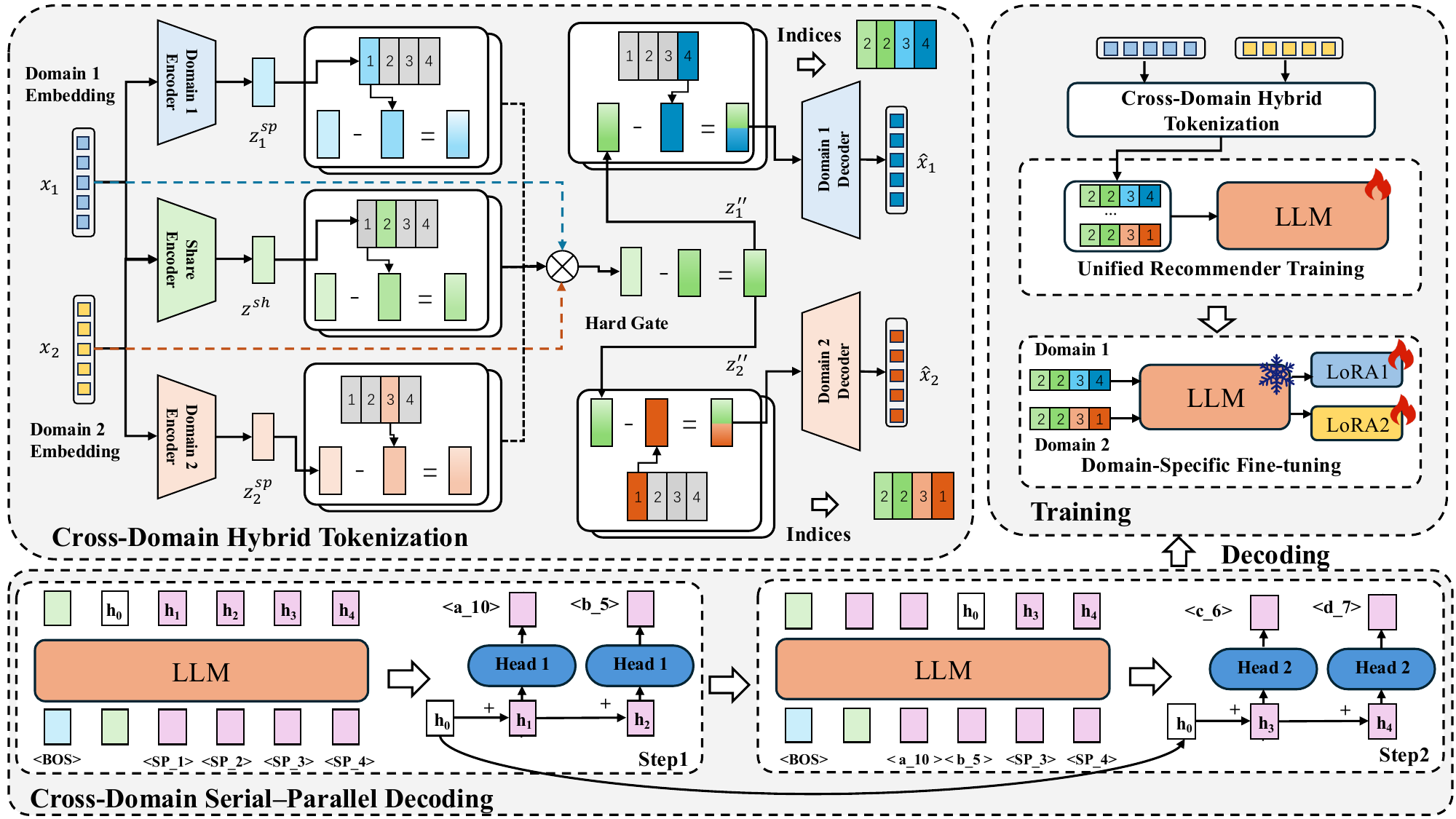}
    \caption{Overview of the proposed GenCDSR framework. 
    % GenCDSR consists of a Cross-Domain Hybrid Tokenization module and a Cross-Domain Serial-Parallel Decoding module, which jointly enable effective cross-domain collaboration modeling and efficient generative recommendation.
    }
    \label{fig:framework}
\end{figure*}

Existing generative cross-domain sequential recommendation methods~\cite{hu2026ids,jin2025generative} fail to disentangle shared and domain-specific factors, relying on fully shared or purely semantic representations. Furthermore, their token-by-token autoregressive decoding limits inference efficiency. To address these issues, we propose GenCDSR, an efficient generative framework comprising core components: Cross-Domain Hybrid Tokenization and Cross-Domain Serial-Parallel Decoding. The framework is shown in Figure~\ref{fig:framework}.

\subsection{Cross-Domain Hybrid Tokenization}

Cross-domain sequential recommendation requires balancing shared semantic commonalities with domain-specific characteristics. However, existing generative methods either adopt fully shared representation spaces~\cite{jin2025generative}, which weaken domain-specific features, or rely on semantic representations~\cite{hu2026ids} that overlook cross-domain collaborative signals. To address this issue, we introduce cross-domain hybrid tokenization (Figure~\ref{fig:framework}, upper left), which employs a multi-tower architecture to capture both factors. It follows a two-stage hierarchy: Stage~1 quantizes shared and specific representations into a length-$L_1$ SID segment, while Stage~2 generates a length-$L_2$ domain-specific segment for finer details. Each item $v_i$ is thus represented by a hierarchical SID sequence $v_i=[t_1,\ldots,t_L]$, where $L=L_1+L_2$ and $t_i$ denotes the SID from the $i$-th-level codebook.

% To address this issue, we introduce a cross-domain hybrid tokenization module (Figure~\ref{fig:framework}, upper left) based on a multi-tower architecture to capture both collaborative signals and domain-specific distinctions. Tokenization follows a two-stage hierarchy: Stage~1 quantizes shared and specific representations into a length-$L_1$ SID segment, and Stage~2 applies fully domain-specific tokenization to produce a length-$L_2$ segment for finer details. Each item $v_i$ is thus represented as a hierarchical SID sequence $v_i=[t_1,\ldots,t_L]$, where $L=L_1+L_2$ and $t_\ell$ denotes the codeword from the $\ell$-th level codebook.

\subsubsection{Stage 1: Shared–Specific Tokenization}
%尽管不同领域在数据分布与行为模式上存在显著差异，它们仍然共享相对稳定的高层语义结构与行为规律。若在 Token 化阶段对各领域完全独立建模，容易导致语义空间的割裂，从而削弱跨领域泛化能力并限制协同建模效果。因此，在第一层中，我们引入了share–specific 的编码与量化机制，以在保留跨域共享语义结构的同时，为后续领域差异建模奠定统一的语义基础。
% Although different domains exhibit substantial differences in data distributions and behavioral patterns, they still share relatively stable high-level semantic structures and behavioral regularities. Modeling each domain independently during the tokenization stage can easily lead to fragmentation in the semantic space, thereby weakening cross-domain generalization and limiting the effectiveness of collaborative modeling. Therefore, in the first stage, building upon the RQ-VAE tokenization structure, we introduce a Share-Specific Cross-Domain Tokenization mechanism to preserve shared semantic structures across domains while establishing a unified semantic foundation for subsequent modeling of domain-specific variations.

% Although different domains exhibit distinct data distributions, they share stable high-level semantics and behavioral patterns. Independent tokenization fragments this space, weakening cross-domain generalization and collaborative modeling. Therefore, our first stage extends the RQ-VAE structure with a Share-Specific Tokenization mechanism. This preserves shared and domain-specific semantics and establishes a unified foundation for subsequent domain-specific modeling.

Although different domains exhibit distinct data distributions, they still share relatively stable high-level semantics and behavioral regularities. Therefore, tokenizing each domain independently can fragment the semantic space, thereby weakening cross-domain generalization and limiting collaborative modeling~\cite{rajput2023recommender,hu2026ids}. To address this, our first stage extends the RQ-VAE tokenization framework with a Shared-Specific Tokenization (SST) mechanism, which jointly preserves shared semantics and domain-specific signals and establishes a unified foundation for subsequent domain-specific modeling.

% 具体而言，我们在 RQ‑VAE 的 Token 化结构基础上，引入 share–specific 的跨域量化机制。对于每个领域，分别构建领域特定的编码器与残差量化模块，同时引入一组跨领域共享的编码器与残差量化模块，从而在统一框架下同时建模跨域共性语义与领域特定特征。
% 给定物品的连续语义表示 x∈RD\mathbf{x} \in \mathbb{R}^Dx∈RD，其潜在表示可以通过共享编码器或对应领域的特定编码器获得：
% Specifically, for each domain, a domain-specific encoder and residual Tokenization module are constructed, while a shared encoder and residual Tokenization module are introduced across domains to provide a unified semantic backbone.
% Given an item belonging to domain $d \in \{A, B\}$, with its semantic representation $\mathbf{x}_d$ extracted by pre-trained models, such as LLaMA~\cite{touvron2023llama} and Qwen~\cite{bai2023qwen}, we first map the input into latent semantic spaces using both a shared encoder and a domain-specific encoder:

Specifically, we construct both shared and domain-specific encoders and residual tokenization modules to provide a unified semantic backbone alongside distinct domain features. For an item in domain $d \in \{A, B\}$ with semantic representation $\mathbf{x}_d$, extracted via pre-trained models like LLaMA-7B~\cite{touvron2023llama} or Qwen2.5-7B~\cite{bai2023qwen}, we first map it into latent semantic spaces using both encoders:
\begin{equation}
\label{equ:SSCDT_1_enc}
\mathbf{z}^{\mathrm{sh}} = \mathrm{Enc}_{\mathrm{sh}}(\mathbf{x}_d), \quad
\mathbf{z}^{\mathrm{sp}}_{d} = \mathrm{Enc}_{d}(\mathbf{x}_d),
\end{equation}
where $\mathrm{Enc}_{\mathrm{sh}}(\cdot)$ denotes the shared encoder and $\mathrm{Enc}_{d}(\cdot)$ denotes the domain-specific encoder for domain $d$. The shared encoder is applied to item embeddings from all domains, whereas each domain-specific encoder is only applied to items within its domain.
% 相应地，共享潜在表示 zshare\mathbf{z}^{\mathrm{share}}zshare 与领域特定潜在表示 zdspec\mathbf{z}^{\mathrm{spec}}_{d}zdspec 分别输入至独立的残差量化模块，通过多级残差量化过程映射为离散语义 ID。

Correspondingly, the shared latent representation $\mathbf{z}^{\mathrm{sh}}$ and the domain-specific latent representation $\mathbf{z}^{\mathrm{sp}}_{d}$ are fed into two independent residual quantization modules. Each representation is then quantized into discrete SIDs via an $L_1$-level residual quantization process, producing $\mathbf{z}'^{\mathrm{sh}}$ and $\mathbf{z}'^{\mathrm{sp}}_{d}$ as the discrete latent representations from the shared and domain-specific branches.

% Correspondingly, $\mathbf{z}^{\mathrm{share}}$ and $\mathbf{z}^{\mathrm{spec}}_{d}$ are quantized by two independent $L_1$-level residual quantizers, yielding the discrete latents $\mathbf{z}'^{\mathrm{share}}$ and $\mathbf{z}'^{\mathrm{spec}}_{d}$, respectively.

%为自适应的选择共享分支与领域特定分支的路径，我们引入 Gumbel-Softmax 机制，对共享token化分支与领域特定token化分支进行自适应选择；具体而言，对于给定物品，模型通过 Gumbel-Softmax 采样得到近似离散的选择权重，最终物品的量化潜在表示由共享token化结果与领域特定token化结果加权组合而成：
% To adaptively select between the shared and domain-specific branches, we introduce a Gumbel-Softmax mechanism to dynamically route each item between the shared tokenization branch and the domain-specific tokenization branch in an end-to-end manner. Specifically, for an item from domain $d$, the router samples near-discrete routing weights via Gumbel-Softmax, and constructs the fused quantized representation as a weighted combination of the shared and domain-specific outputs:

To adaptively choose between the shared and domain-specific branches, we use a Gumbel-Softmax router to sample near-discrete routing weights and fuse the corresponding quantized outputs:
\begin{align}
\label{equ:SSCDT_1_gumble}
&\boldsymbol{\pi}_d=f\!\left(\mathbf{x}_d\right),\\
&g^{\mathrm{sh}}, g^{\mathrm{sp}}_d = \mathrm{GumbelSoftmax}(\boldsymbol{\pi}_d;\tau),
\\
&\mathbf{z}'_d
=
g^{\mathrm{sh}} \cdot \mathbf{z}'^{\mathrm{sh}}
+
g^{\mathrm{sp}}_d \cdot \mathbf{z}'^{\mathrm{sp}}_d,
\end{align}
where $f(\cdot)$ is an MLP and $\tau$ is the temperature.

%为了保留RQ-VAE过程表示未捕获的信息，我们通过从原始潜在特征中减去Gumbel-Softmax融合量化输出来计算残差，得到的最终表示传递到下层。

% To preserve the information not captured by the representations generated in the RQ-VAE, we compute the residual by subtracting the Gumbel-Softmax-fused output $\mathbf{z}'$ from the original latent representation, and the resulting final representation of the Stage1 is then fed to the tokenization Stage2 and we get $[t_1,\ldots,t_{L_1}]$. The final representation:

To preserve information not captured by the Stage~1 quantized representations, we compute a residual by subtracting the Gumbel-Softmax fused quantized output $\mathbf{z}'_d$ from the corresponding fused continuous latent feature. This residual, denoted as $\mathbf{z}''_d$, serves as the input to Stage~2 for further domain-specific tokenization. At this point, Stage~1 has already produced the final $L_1$ SID tokens $[t_1,\ldots,t_{L_1}]$. 
Specifically, the Stage~2 input is defined as:
\begin{equation}
\label{equ:SSCDT_1_residual}
\mathbf{z}''_d=\Big(g^{\mathrm{sh}} \cdot \mathbf{z}^{\mathrm{sh}}
+
g^{\mathrm{sp}}_d \cdot \mathbf{z}^{\mathrm{sp}}_d\Big)-\mathbf{z'}_d.
\end{equation}

% \begin{equation}
% \label{equ:SSCDT_1_residual}
% \mathbf{z}''_d=
% \Big(g^{\mathrm{share}}_d \cdot \mathbf{z}^{\mathrm{share}}_d
% +
% g^{\mathrm{spec}}_d \cdot \mathbf{z}^{\mathrm{spec}}_d\Big)
% -\mathbf{z}'_d.
% \end{equation}
\subsubsection{Stage 2: Fine-Grained Specific Tokenization}

%在实际推荐场景中，不同领域往往具有显著差异的物品分布。若在低层表示中仍使用共享 codebook，容易对领域特定模式施加过强约束，甚至导致语义混淆。因此，在第二层中采用完全领域（场景）特定的 RQ-VAE 量化模块，以更充分地刻画各域的细粒度差异。随后，多层 RQ-VAE 量化得到的表征被输入到对应领域的解码器中进行重建：
% In practical recommendation systems, different domains often exhibit substantially different item distributions. If a shared codebook is still used for low-level representations, domain-specific patterns may be overly constrained, which can even lead to semantic ambiguity. Therefore, in the second step, we employ fully domain-specific RQ tokenization modules to better capture fine-grained, domain-dependent variations. Starting from the Stage~2 input $\mathbf{z}''_d$, we apply an $L_2$-level residual quantization to obtain the domain-specific quantized representation $\hat{\mathbf{z}}_{d}$, which is then fed into the corresponding domain decoder for reconstruction:
% \begin{equation}
% \label{equ:SSCDT_2_decoder}
% \hat{\mathbf{x}}_{d} = \mathrm{Dec}_{d}\!\left(\hat{\mathbf{z}}_{d}\right),
% \end{equation}
% where $\mathrm{Dec}_{d}(\cdot)$ is the decoder associated with domain $d$.

In practical recommendation systems, different domains often exhibit substantially different item distributions and interaction patterns. If a shared codebook is still used for low-level representations, domain-specific patterns may be overly constrained, which can even lead to semantic ambiguity~\cite{jin2025generative,hu2026ids}. Therefore, in Stage~2, we employ fully domain-specific RQ tokenization named Fine-Grained Specific Tokenization (FGST) modules to better capture fine-grained, domain-dependent variations. Starting from the Stage~2 input $\mathbf{z}''_d$, we apply an $L_2$-level residual quantization to obtain the domain-specific quantized representation $\hat{\mathbf{z}}_{d}$, which is then fed into the corresponding domain decoder for reconstruction:
% In practical recommendation systems, different domains often exhibit substantially different item distributions. If a shared codebook is still used for low-level representations, domain-specific patterns may be overly constrained, which can even lead to semantic ambiguity. Therefore, in Stage~2, we employ fully domain-specific RQ tokenization modules to better capture fine-grained, domain-dependent variations. Starting from the Stage~2 input $\mathbf{z}''_d$, we apply an $L_2$-level residual quantization to obtain the domain-specific quantized representation $\hat{\mathbf{z}}_{d}$, which is then fed into the corresponding domain decoder for reconstruction:
\begin{equation}
\label{equ:SSCDT_2_decoder}
\hat{\mathbf{x}}_{d} = \mathrm{Dec}_{d}\!\left(\hat{\mathbf{z}}_{d}\right),
\end{equation}
where $\mathrm{Dec}_{d}(\cdot)$ is the decoder associated with domain $d$. 
After Stage~2, the remaining $L_2$ SID tokens are determined and combined with the $L_1$ tokens from Stage~1, yielding the complete SID sequence $[t_1,\ldots,t_L]$ with $L=L_1+L_2$.
% \zc{
% \subsubsection{Stage2: Fine-Grained Specific Tokenization}

% }

% The loss for residual quantization is defined as:
% \begin{align}
% \label{equ:SSCDT_2_RQ}
% &\mathcal{L}^{\mathrm{share}}_{\mathrm{RQ}}
% =
% \sum_{l=1}^{L_{\mathrm{share}}}
% \Big(
% \big\| \mathrm{sg}(\mathbf{r}^{(l-1)}_{\mathrm{share}}) - \mathbf{c}^{(l)}_{\mathrm{share},\,s^{(l)}_{\mathrm{share}}} \big\|_2^2
% +
% \beta \big\| \mathbf{r}^{(l-1)}_{\mathrm{share}} - \mathrm{sg}(\mathbf{c}^{(l)}_{\mathrm{share},\,s^{(l)}_{\mathrm{share}}}) \big\|_2^2
% \Big), \nonumber\\
% &\mathcal{L}^{\mathrm{spec},\mathcal{P}}_{\mathrm{RQ}}
% =
% \sum_{l=1}^{L_{\mathrm{spec}}}
% \Big(
% \big\| \mathrm{sg}(\mathbf{r}^{(l-1)}_{\mathrm{spec},\mathcal{P}}) - \mathbf{c}^{(l)}_{\mathrm{spec},\mathcal{P},\,s^{(l)}_{\mathrm{spec},\mathcal{P}}} \big\|_2^2
% +
% \beta \big\| \mathbf{r}^{(l-1)}_{\mathrm{spec},\mathcal{P}} - \mathrm{sg}(\mathbf{c}^{(l)}_{\mathrm{spec},\mathcal{P},\,s^{(l)}_{\mathrm{spec},\mathcal{P}}}) \big\|_2^2
% \Big), \nonumber\\
% &\mathcal{L}^{\mathrm{spec},\mathcal{Q}}_{\mathrm{RQ}}
% =
% \sum_{l=1}^{L_{\mathrm{spec}}}
% \Big(
% \big\| \mathrm{sg}(\mathbf{r}^{(l-1)}_{\mathrm{spec},\mathcal{Q}}) - \mathbf{c}^{(l)}_{\mathrm{spec},\mathcal{Q},\,s^{(l)}_{\mathrm{spec},\mathcal{Q}}} \big\|_2^2
% +
% \beta \big\| \mathbf{r}^{(l-1)}_{\mathrm{spec},\mathcal{Q}} - \mathrm{sg}(\mathbf{c}^{(l)}_{\mathrm{spec},\mathcal{Q},\,s^{(l)}_{\mathrm{spec},\mathcal{Q}}}) \big\|_2^2
% \Big).
% \end{align}

% where $\mathrm{sg}(\cdot)$ denotes the stop-gradient operator and $\beta$ is the coefficient balancing codebook learning and encoder updates. 
The total loss for tokenization can be formulated as:
\begin{align}
\label{equ:SSCDT_2_total_loss}
&\mathcal{L}_{\mathrm{recon}} = 
\sum_{d \in \{A,B\}} \big\| \mathbf{x}_{d} - \hat{\mathbf{x}}_{d} \big\|_2^2,\\
&\mathcal{L}_{\mathrm{RQ}} =
\mathcal{L}^{\mathrm{sh}}_{\mathrm{RQ}}
+
\mathcal{L}^{\mathrm{sp},A}_{\mathrm{RQ}}
+
\mathcal{L}^{\mathrm{sp},B}_{\mathrm{RQ}},\\
&\mathcal{L} = \mathcal{L}_{\mathrm{recon}} + \mathcal{L}_{\mathrm{RQ}}.
\end{align}
where $\mathcal{L}^{\mathrm{sh}}_{\mathrm{RQ}}$ is the RQ loss of the shared tokenization, while $\mathcal{L}^{\mathrm{sp},A}_{\mathrm{RQ}}$ and $ $$\mathcal{L}^{\mathrm{sp},B}_{\mathrm{RQ}}$ are the RQ losses of the domain-specific tokenization for domains $A$ and $B$.
\subsection{Cross-Domain Serial-Parallel Decoding}
%在生成式跨域推荐中，现有方法多采用逐 token 的串行解码，使推理时延随场景数量增加而显著上升，成为模型实际部署的主要瓶颈之一。与此同时，这种纯串行的解码范式往往难以充分利用我们在 tokenization 阶段所构造的两阶段语义 ID 结构，从而限制了跨域语义对齐与协同信息的有效传播与表达。因此，如何在保证生成质量的前提下，设计一种高效且可扩展的多场景解码机制，并更好地匹配两阶段语义标识的生成结构，是生成式跨域推荐面临的另一项核心挑战。为此，本文提出 Cross-Domain Hybrid Serial–Parallel Decoding 解码策略。
% In generative cross-domain recommendation, most existing methods~\cite{hu2026ids,jin2025generative} rely on token-by-token serial decoding, which causes inference latency to increase substantially and becomes a major bottleneck for practical deployment. Meanwhile, such a purely serial decoding paradigm often fails to fully exploit the two-stage semantic ID structure constructed during tokenization, thereby limiting effective propagation and utilization of cross-domain collaborative signals. Therefore, designing an efficient and scalable multi-domain decoding mechanism that preserves generation quality while better matching the two-stage identifier generation process remains another core challenge in generative cross-domain recommendation. To this end, we propose a Cross-Domain Hybrid Serial-Parallel Decoding strategy.

In generative cross-domain sequential recommendation, existing methods~\cite{hu2026ids,jin2025generative} typically rely on token-by-token serial decoding. This creates a severe latency bottleneck for practical deployment and underutilizes the two-stage SID structure, thereby limiting the propagation of collaborative signals. To achieve efficient and scalable generation that better matches this hierarchical identifier design, we propose a cross-domain serial-parallel decoding strategy, as shown at the bottom of Figure~\ref{fig:framework}.

%我们设计了一种两阶段解码方案，使其既能适配跨域语义码本，又能在保证推理效率的同时维持生成效果。具体而言，对于每个训练样本 {(q,x),y}\{(q,\mathbf{x}),\mathbf{y}\}{(q,x),y}，经过Cross-Domain Hybrid Tokenization后，将目标语义 ID 表示为由 L=L1+L2 个特殊令牌组成的序列，即 y=[t1y,t2y,…,tLy]\mathbf{y}=[t_1^{y},t_2^{y},\ldots,t_L^{y}]y=[t1y​,t2y​,…,tLy​]。在构造输入提示时，用 L个占位符令牌替换真实目标序列，从而在一次前向传播中预计算未来令牌位置对应的隐状态。设 h0,h1,…,hLh_0,h_1,\ldots,h_Lh0​,h1​,…,hL​ 为大语言模型（LLM）最后一层输出的隐状态，其中 h0h_0h0​ 对应前缀部分，h1∼hLh_1\sim h_Lh1​∼hL​ 分别对应 LLL 个占位符令牌。基于这些隐状态，解码过程进一步划分为两个阶段进行：

After cross-domain hybrid tokenization, the target SID of $v_{n+1}$ is represented as a length-$L$ sequence of special tokens, i.e., $v_{n+1}=[t_1,t_2,\ldots,t_L]$. During prompting, the ground-truth target SID tokens are replaced with $L$ placeholder tokens $[\texttt{<SP\_1>}, \ldots, \texttt{<SP\_L>}]$, which enables a single forward pass to pre-compute the hidden states for all future token positions in a prefix-conditioned manner. 
% Let $h_0,h_1,\ldots,h_L$ denote the last-layer hidden states of the large language model, where $h_0$ corresponds to the prefix context and $h_1\sim h_L$ correspond to the $L$ placeholder positions, respectively. Based on these hidden states, the decoding procedure is further performed in two steps. We define an evolving context state $s_l$ to summarize the decoding history, which is initialized by the prefix representation:
Let $h^{1}_0,h^{1}_1,\ldots,h^{1}_{L_1}$ denote the last-layer hidden states from the first LLM call, which correspond to the prefix $h^{1}_0$ and the $L_1$ placeholder positions $h^{1}_1\sim h^{1}_{L_1}$. We then define an evolving context state $s_l$ initialized by the prefix:
\begin{equation}
s_1 = h_0 .
\label{equ:CDHSPD_s1}
\end{equation}

% For $l_1=1,\ldots,L_1$, the first-stage head $\mathrm{Head}_1$ predicts the $l_1$-th semantic ID's probability distribution $\mathbf{p}_{l_1}$ and updates the state by summing the embedding $\mathbf{e}_{l_1}$ of the newly selected token \zc{unclear for $\mathrm{Head}_1(h_{l_1}, s_{l_1})$}:

% For $l=1,\ldots,L_1$, the Step~1 head produces the token distribution at position $l$:
% \begin{equation}
% % \mathbf{p}_{l_1} = \mathrm{Head}_1(h_{l_1}, s_{l_1}),\quad
% % s_{{l_1}+1} = s_{l_1} + \mathbf{e}_{l_1}.
% \mathbf{p}_l = \mathrm{softmax}\big(\mathrm{Head}_1(h^{1}_l, s_l)\big),
% \quad 
% s_{l+1} = s_l + \mathbf{e}_l,
% \label{equ:CDHSPD_step1}
% \end{equation}
% where $\mathbf{e}_l=\mathrm{Emb}_l(t_l)$ is the embedding of the selected token $t_l$.

For $l=1,\ldots,L_1$, the Step~1 head produces the token distribution at position $l$ and updates the context state using the embedding $\mathbf{e}_l=\mathrm{Emb}_l(t_l)$ of the selected token $t_l$:
\begin{equation}
\mathbf{p}_l^{(1)} = \mathrm{softmax}\big(\mathrm{Head}_1(h^{1}_l, s_l)\big),
\quad 
s_{l+1} = s_l + \mathbf{e}_l.
\label{equ:CDHSPD_step1}
\end{equation}

% Next, we invoke the LLM once more to perform the second-stage decoding pass. Let $s_{L_1+1}$ denote the carried state after Stage~1, which is then passed to Stage~2. For $l_2 = 1,\ldots,L_2$, the second-stage head $\mathrm{Head}_2$ generates the remaining semantic ID tokens as follows:
Next, we invoke the LLM once more by filling the first $L_1$ placeholders with the Step~1 tokens, obtaining updated hidden states $h^{2}_0,h^{2}_1,\ldots,h^{2}_L$. Let $s_{L_1+1}$ be the carried state after Step~1. For each remaining position $l\in\{L_1+1,\ldots,L\}$, the Step~2 head predicts in parallel conditioned on the carried state:
\begin{equation}
% \mathbf{p}_{l_2} = \mathrm{Head}_2(h_{l_2}, s_{l_2}),\quad
% s_{{l_2}+1} = s_{l_2} + \mathbf{e}_{l_2}.
\mathbf{p}_l^{(2)} = \mathrm{softmax}\big(\mathrm{Head}_2(h^{2}_l, s_{L_1+1})\big),\quad s_{l+1} = s_l + \mathbf{e}_l.
\label{equ:CDHSPD_stage2}
\end{equation}

% Let $h^{(1)}_0,h^{(1)}_1,\ldots,h^{(1)}_L$ denote the last-layer hidden states from the first LLM call, corresponding to the prefix ($h^{(1)}_0$) and the $L$ placeholder positions ($h^{(1)}_1\sim h^{(1)}_L$). We define an evolving context state $s_\ell$ initialized by the prefix:
% \begin{equation}
% s_1 = h^{(1)}_0 .
% \label{eq:cdhspd_init}
% \end{equation}

% \paragraph{Stage~1 (serial; $\ell=1,\ldots,L_1$).}
% For $\ell=1,\ldots,L_1$, the first-stage head produces the token distribution at position $\ell$:
% \begin{equation}
% \mathbf{p}_\ell = \mathrm{softmax}\!\big(\mathrm{Head}_1(h^{(1)}_\ell, s_\ell)\big),
% \quad 
% s_{\ell+1} = s_\ell + \mathbf{e}_\ell,
% \label{eq:cdhspd_stage1}
% \end{equation}
% where $\mathbf{e}_\ell=\mathrm{Emb}_\ell(t_\ell)$ is the embedding of the selected token $t_\ell$ (teacher-forced during training).

% \paragraph{Second LLM call.}
% Next, we invoke the LLM once more by filling the first $L_1$ placeholders with the Stage~1 tokens, obtaining updated hidden states $h^{(2)}_0,h^{(2)}_1,\ldots,h^{(2)}_L$.

% \paragraph{Stage~2 (parallel; $\ell=L_1+1,\ldots,L$).}
% Let $s_{L_1+1}$ be the carried state after Stage~1. For each remaining position $\ell\in\{L_1+1,\ldots,L\}$, the second-stage head predicts in parallel conditioned on the carried state:
% \begin{equation}
% \mathbf{p}_\ell = \mathrm{softmax}\!\big(\mathrm{Head}_2(h^{(2)}_\ell, s_{L_1+1})\big).
% \label{eq:cdhspd_stage2}
% \end{equation}

\subsection{Training and Inference}

%在本节中，我们介绍模型的整体训练与推理过程。遵循先前研究，我们采用两阶段训练策略，以在跨域知识共享与领域差异适配之间取得平衡。第一阶段进行Unified Recommender Training：使用合并领域数据，从而增强模型对跨域模式的理解能力与知识迁移能力。模型通过最大化目标语义 ID 序列的逐 token 条件似然进行优化：
% In this subsection, we describe the overall training and inference procedures of our framework. Following prior work~\cite{liu2025bridge}, we adopt a two-stage training strategy to balance cross-domain knowledge sharing and domain-specific adaptation.

% In this subsection, we describe the overall training and inference procedures of our framework. Following common practice in generative cross-domain sequential recommendation~\cite{liu2025bridge}, we adopt a two-phase training schedule to balance cross-domain knowledge sharing and domain-specific adaptation.
% Following common practice in generative cross-domain sequential recommendation~\cite{liu2025bridge}, this subsection outlines our unified training and inference procedures, balancing cross-domain knowledge sharing and domain-specific adaptation (Figure~\ref{fig:framework}, upper right).
\subsubsection{Unified Recommender Training}
Firstly, we conduct Unified Recommender Training on the merged cross-domain data $\bar{S}_u$ with user $u\in\mathcal{U}$ as Equation~\eqref{eq:problem_behavior} to enhance the model's ability to capture transferable cross-domain patterns. The unified objective maximizes the token-level cross-entropy over all SID positions:
\begin{equation}
\label{eq:unified-train}
% \max_{\Phi}\ \sum_{\bar{S}_u}\ (\sum_{{l_1}=1}^{L_1}\log \mathbf{p}_{l_1}+\sum_{{l_2}=1}^{L_2}\log \mathbf{p}_{l_2}),
\max_{\Phi}\ \sum_{u\in\mathcal{U}}
\left(
\sum_{l=1}^{L_1}\log \mathbf{p}^{(\Phi,1)}_l(t_l)
+
\sum_{l=L_1+1}^{L}\log \mathbf{p}^{(\Phi,2)}_l(t_l)
\right),
\end{equation}
% where $\Phi$ denotes the backbone parameters, $\mathbf{v}_{n+1}=[t_1,\ldots,t_L]$ is the target semantic ID sequence with $L=L_1+L_2$, and % $\mathbf{p}^{1}_l(\cdot)$/$\mathbf{p}^{2}_l(\cdot)$ are the token distributions produced by $\mathrm{Head}_1$ and $\mathrm{Head}_2$, respectively.
% $\mathbf{p}^{(\Phi,1)}_\ell(\cdot)$ and $\mathbf{p}^{(\Phi,2)}_\ell(\cdot)$ denote the token distributions produced under backbone parameters $\Phi$ in Step~1 and Step~2 decoding, respectively.
where $\Phi$ denotes the backbone parameters, and $v_{n+1}=[t_1,\ldots,t_L]$ is the target SID sequence with $L=L_1+L_2$. Moreover, $\mathbf{p}^{(\Phi,1)}_l(\cdot)$ and $\mathbf{p}^{(\Phi,2)}_l(\cdot)$ denote the token distributions produced under $\Phi$ in Step~1 and Step~2 decoding, respectively.

\subsubsection{Domain-Specific Fine-tuning}
%第二阶段为更好地适配各领域的分布差异，同时避免覆盖第一阶段学到的跨域共享知识，我们冻结主干模型参数，并为每个领域 单独引入并微调一个轻量级 LoRA 模块。对应的领域特定优化目标为：
% Subsequently, to better accommodate domain-specific data characteristics while avoiding overwriting the cross-domain knowledge learned in Unified Recommender Training, we freeze the backbone parameters $\Phi$ and fine-tune a lightweight domain-specific LoRA~\cite{hu2022lora} module $\Theta_d$ for each domain $d$ with the data $S_u^{d}$ as Equation~\eqref{eq:problem_user}. The domain-specific objective is:

Subsequently, to better accommodate domain-specific data characteristics while avoiding overwriting the cross-domain knowledge learned during Unified Recommender Training, we freeze the backbone parameters $\Phi$ and fine-tune a lightweight, domain-specific LoRA~\cite{hu2022lora} adapter $\Theta_d$ for each domain $d$ using the corresponding data $S_u^{d}$ in Equation~\eqref{eq:problem_user}. The resulting domain-specific optimization objective is:
\begin{equation}
% \max_{\Theta_d}\ \sum_{S_u^{d}}\ (\sum_{{l_1}=1}^{L_1}\log \mathbf{p}_{l_1}+\sum_{{l_2}=1}^{L_2}\log \mathbf{p}_{l_2}).
\max_{\Theta_d}\ \sum_{u\in\mathcal{U}}
\left(
\sum_{l=1}^{L_1}\log \mathbf{p}^{(\Phi+\Theta_d,1)}_l(t_l)
+
\sum_{l=L_1+1}^{L}\log \mathbf{p}^{(\Phi+\Theta_d,2)}_l(t_l)
\right).
\label{eq:lora-ft}
\end{equation}
% where $\mathbf{p}^{(\Phi+\Theta_d,1)}_\ell(\cdot)$ and $\mathbf{p}^{(\Phi+\Theta_d,2)}_\ell(\cdot)$ denote the token distributions computed under the backbone $\Phi$ with the domain adapter $\Theta_d$ activated, in the first and second LLM calls, respectively.
\subsubsection{Inference}
%在推理阶段，模型自回归生成语义 ID 序列并据此检索对应物品。为确保生成的标识符始终对应物品库 中的有效物品，我们沿用已有工作并在由中物品语义 ID 构建的前缀树（Trie）上执行约束生成，从而将搜索空间限制在可行前缀集合中：

During inference, we activate the domain-specific adapter $\Theta_d$ on top of the backbone $\Phi$ and perform trie-constrained serial-parallel decoding to generate a SID sequence, which is then mapped to the corresponding item. To ensure that the generated identifier corresponds to a valid item in $\mathcal{I}_d$, we follow existing work~\cite{hu2026ids,jin2025generative,rajput2023recommender} and restrict decoding to the feasible set $\mathcal{T}_d$ induced by the Trie~\cite{de1959file}. The inference objective is:
\begin{equation}
% \hat{v}_{n}
% =
% \argmax_{\mathbf{t_l}\in \mathcal{I}_d}
% \Bigg(
% \sum_{l_1=1}^{L_1}\log \mathbf{p}_{l_1}^{\Phi+\Theta_d}(t_{l_1})
% +
% \sum_{l_2=1}^{L_2}\log \mathbf{p}_{l_2}^{\Phi+\Theta_d}(t_{l_2})
% \Bigg).         
\hat{v}_{n+1}
=
\argmax_{v_{n+1}\in\mathcal{T}_d}
\left(
\sum_{l=1}^{L_1}\log \mathbf{p}^{(\Phi+\Theta_d,1)}_l(t_l)
+
\sum_{l=L_1+1}^{L}\log \mathbf{p}^{(\Phi+\Theta_d,2)}_l(t_l)
\right).
\label{eq:trie}
\end{equation}

\section{Experiments}
In this section, we conduct extensive experiments on three public datasets and answer the following research questions (RQs):

% GenCDSR 在跨域数据集上是否优于强基线方法？
%混合串并行解码能否在不同骨干下显著降低延迟且不牺牲效果？
%Share–Specific Tokenization与Fine-Grained Specific Tokenization的tokenization 各自贡献是什么，二者是否缺一不可？
%Share–Specific Tokenization与Fine-Grained Specific Tokenization分配比例会对我们的模型有怎样的表现
% 本文 tokenization 是否能保留更全面的语义特征？

% \begin{itemize}[leftmargin=*]
% \item \textbf{RQ1:}
% \end{itemize}

% \begin{itemize}[leftmargin=*]
%   \item \textbf{RQ1:} Does GenCDSR consistently outperform strong baselines on cross-domain datasets?
%   \item \textbf{RQ2:} Can Cross-Domain Serial–Parallel Decoding significantly reduce inference latency under different backbones without sacrificing recommendation quality?
%   \item \textbf{RQ3:} What are the respective contributions of Share-Specific Tokenization and Fine-Grained Specific Tokenization, and are both components necessary?
%   \item \textbf{RQ4:} How does the allocation ratio between Share-Specific Tokenization and Fine-Grained Specific Tokenization codebooks affect performance, and is there a stable optimal range?
%   \item \textbf{RQ5:} Does our tokenization achieve higher feature fidelity for semantic IDs?
% \end{itemize}

\begin{itemize}[leftmargin=*]
  \item \textbf{RQ1:} What is the performance of GenCDSR compared with SOTA baseline methods?%Does GenCDSR consistently outperform strong baselines on cross-domain datasets?
  \item \textbf{RQ2:} Can cross-domain serial-parallel decoding reduce inference latency without sacrificing accuracy? %across backbones
  \item \textbf{RQ3:} What are the effects of shared-specific tokenization and fine-grained specific tokenization? %, and are both necessary?
  \item \textbf{RQ4:} How does the cross-domain hybrid tokenization contribute to improving the recommendation performance? %Does our tokenization improve feature fidelity of SIDs?
  \item \textbf{RQ5:} How does the allocation of shared-specific tokenization and fine-grained specific tokenization affect the  performance? %, and is there a stable optimal range?
\end{itemize}

\subsection{Experimental Settings}
\subsubsection{Datasets}

Following prior CDSR studies~\cite{liu2025llm,liu2025bridge}, we evaluate our method on three public cross-domain datasets: Clothing--Sports, Electronics--Phone, and Book--Movie. The first two are collected from Amazon\footnote{\url{https://jmcauley.ucsd.edu/data/amazon/index_2014.html}}, while the third is collected from Douban\footnote{\url{https://github.com/fengzhu1/GA-DTCDR/tree/main/Data}}. We filter out users with fewer than five interactions and items with fewer than three interactions in either domain, and chronologically merge cross-domain interactions into mixed behavior sequences for each user. Following the leave-one-out strategy~\cite{zheng2024adapting,rajput2023recommender}, we use the last two interactions for validation and testing, respectively. Table~\ref{tab:dataset_stats} summarizes the dataset statistics.

\begin{table}[t]
\setlength\abovecaptionskip{0.2\baselineskip}
\setlength\belowcaptionskip{0.2\baselineskip}
\centering
\caption{Dataset statistics, including the number of users, number of items, sparsity, the number of overlapped users, and the average interaction length.}
\label{tab:dataset_stats}
\begin{tabular}{lccc cc}
\toprule
\textbf{Dataset} & \textbf{Users} & \textbf{Items} & \textbf{Sparsity} & \textbf{Overlap} & \textbf{Avg.len} \\
\midrule
Clothing  & 9,933  & 3,278  & 99.70\% & \multirow{2}{*}{3,962} & \multirow{2}{*}{10.71} \\
Sports  & 4,263  & 1,021  & 99.04\% &                       &                       \\
\midrule
Electronics & 20,728 & 10,492 & 99.93\% & \multirow{2}{*}{11,698} & \multirow{2}{*}{8.30} \\
Phone      & 11,762 & 2,246  & 99.88\% &                        &                       \\
\midrule
Book       & 1,381  & 12,426 & 99.88\% & \multirow{2}{*}{1,265}  & \multirow{2}{*}{68.30} \\
Movie      & 2,213  & 16,537 & 99.92\% &                        &                       \\
\bottomrule
\end{tabular}
\end{table}
% 为验证所提出方法的有效性，本文选取具有代表性的基线模型进行对比实验，涵盖两个类别：单域序列推荐（Single-domain Sequential Recommendation, SDSR） 和 跨域序列推荐（Cross-domain Sequential Recommendation, CDSR）。并且每个部分又分为了判别式和生成式。
% 单域序列推荐（SDSR）
% 判别式
% 该类方法仅利用单一域内的用户历史行为序列来建模用户偏好，不引入来自其他域的辅助信息。本文采用以下被广泛使用的 SDSR 基线模型：
% GRU4Rec
% GRU4Rec 使用门控循环单元（GRU）对用户行为序列进行建模，以捕捉用户兴趣的动态变化，是较早提出的神经序列推荐方法之一。
% BERT4Rec
% 受 BERT 架构启发，BERT4Rec 通过双向自注意力机制建模用户行为序列中的上下文依赖关系，并采用掩码预测任务进行训练。
% SASRec
% 不同于 BERT4Rec 的双向建模方式，SASRec 采用单向自注意力机制，以刻画序列中的因果关系，更适用于下一物品预测任务。
%生成式
% Tiger
%TIGER 利用 RQ‑VAE 将物品语义信息量化为代码序列，并以此支持基于大语言模型的生成式推荐。
% 跨域序列推荐（CDSR）
%判别式
% 跨域序列推荐方法旨在通过挖掘用户在多个域中的交互行为来缓解数据稀疏问题。本文选取以下具有代表性的 CDSR 基线模型进行对比：
% C2DSR
% C2DSR 利用图神经网络建模不同域中物品之间的关系，并通过图结构实现跨域知识迁移。
% TriCDR
% TriCDR 通过设计跨域对比学习目标，从用户的行为序列中捕捉更加细粒度的兴趣表示，以增强跨域推荐性能。
%LLM4CDSR
%LLM4CDSR 通过引入大语言模型缓解跨域序列推荐中的数据稀疏问题。该方法学习跨域统一的物品语义表示，并结合用户画像建模用户偏好，从而增强跨域推荐能力
%生成式
%Tiger
%我们将TIGER适配多域场景下进行联合训练。
%MusicRec
%MusicRec 利用先share后specific的RQ‑VAE 融合多模态语义与用户协同信号，构造生成式推荐模型，我们这里把他应用到跨域场景中。
%GenCDR
%GenCDR 利用综合语义关联的跨域token化结构， 融合跨域语义信息以支持跨域生成式推荐。
\subsubsection{Baselines}

To evaluate the effectiveness of GenCDSR, we compare it  with representative baseline models, which can be divided into two categories: single-domain sequential recommendation and cross-domain sequential recommendation. 

%For each category, the baseline methods are further divided into \textbf{discriminative} and \textbf{generative} approaches. Additionally, we evaluate several decoding methods in our Component Analysis.

\noindent \paragraph{Single-Domain Sequential Recommendation.}

These methods model user preferences based solely on historical interaction sequences within a single domain, without leveraging auxiliary information from other domains. 
%We adopt the following baselines, in which the first three are discriminative approaches while the last is a generative method.

%\textbf{Discriminative approaches.}

\begin{itemize}[leftmargin=*]
\item \textbf{GRU4Rec}~\cite{hidasi2015session} models user behavior sequences using gated recurrent units to capture dynamic user preferences.
\item \textbf{BERT4Rec}~\cite{sun2019bert4rec} employs bidirectional self-attention to model contextual dependencies in user behavior sequences via a masked prediction objective.
\item \textbf{SASRec}~\cite{kang2018self} adopts unidirectional self-attention to model users' sequential behaviors and capture sequential relationships. %making it suitable for next-item prediction.
\item \textbf{TIGER}~\cite{rajput2023recommender} first utilizes RQ-VAE to obtain SIDs and adopts T5 to generate the SIDs of next item.

%quantize item semantic information into discrete code sequences, enabling LLM-based generative recommendation.
\end{itemize}

%\textbf{Generative approaches:}
%\begin{itemize}[leftmargin=*]
%\end{itemize}

\noindent \paragraph{Cross-Domain Sequential Recommendation.}
CDSR methods aim to alleviate data sparsity by exploiting user interactions across multiple domains. 
%We consider the following representative baselines, in which the first three are discriminative approaches while the last is a representative generative method.

%\textbf{Discriminative approaches:}
\begin{itemize}[leftmargin=*]
\item \textbf{C2DSR}~\cite{cao2022contrastive} models item relationships across domains via graph neural networks to enable cross-domain knowledge transfer.
\item \textbf{TriCDR}~\cite{ma2024triple} captures fine-grained user interests through cross-domain contrastive learning objectives.
\item \textbf{LLM4CDSR}~\cite{liu2025bridge} introduces large language models to learn unified item semantic representations across domains and model user preferences via user profiling.
\item \textbf{GenCDR}~\cite{hu2026ids} leverages a cross-domain tokenization structure with unified semantic associations to support cross-domain generative recommendation.
\end{itemize}

%\textbf{Generative approaches:}
% \begin{itemize}[leftmargin=*]
%   % \item \textbf{TIGER (CDSR Extension)}~\cite{rajput2023recommender} extends TIGER to the multi-domain setting and performs joint training across domains.
%   % \item \textbf{MusicRec}~\cite{zhao2026musicrec} adopts a shared-then-specific RQ-VAE architecture to integrate multimodal semantics and collaborative signals, which is further applied to the cross-domain generative recommendation scenario.
  
% \end{itemize}

\subsubsection{Implementation Details}
% All experiments are conducted on a single NVIDIA RTX PRO 6000 GPU for consistency, and we fix the random seed to 42. 
% For item tokenization, we use LLaMA and Qwen to encode item titles and descriptions into dense embeddings, and then discretize them via an RQ-VAE with a two-stage identifier structure ($L_1=2$ and $L_2=2$). 
% Each level contains 256 code vectors with a dimensionality of 128. 
% We train the tokenization model using AdamW with a learning rate of $1\times10^{-3}$ and a batch size of 1,024. 
% We instantiate our approach on two representative transformer-based generative recommenders, TIGER and Qwen3-0.6B. 
% Since the official TIGER implementation is unavailable, we reimplement it following the original paper; TIGER is optimized with AdamW (learning rate $5\times10^{-4}$, batch size 256) and uses domain-specific LoRA fine-tuning with rank 8 and $\alpha=32$. 
% For Qwen3-0.6B, we optimize with AdamW (learning rate $1\times10^{-4}$, batch size 128) and apply domain-specific LoRA with rank 8 and $\alpha=16$. 
% During inference, we use beam search with a beam size of 20.

Experiments are averaged over 3 runs on 8 NVIDIA RTX PRO 6000 GPUs. %fixed to with the random seed of 42
For item tokenization, we obtain item text embeddings by mean-pooling the last-layer hidden states of LLaMA-7B~\cite{touvron2023llama} and Qwen2.5-7B~\cite{bai2023qwen}, and discretize the embeddings via an RQ-VAE with a two-stage tokenization process ($L_1=2, L_2=2$), where each level has 256 code vectors of 128 dimensions. 
The tokenization model is trained using AdamW~\cite{loshchilovdecoupled} with lr $=1\times10^{-3}$ and batch size $=1{,}024$. 
We instantiate our method on T5~\cite{raffel2020exploring} and Qwen3-0.6B~\cite{yang2025qwen3}. For T5 we use the same %lightweight seq2seq architecture 
setting as TIGER~\cite{rajput2023recommender}, which is optimized using AdamW with lr $=5\times10^{-4}$ and batch size $=256$ and domain-specific LoRA with rank $=8$ and $\alpha=32$. Qwen3-0.6B uses AdamW with lr $=1\times10^{-4}$ and batch size $=128$ and LoRA with rank $=8$ and $\alpha=16$. 
During inference, we apply beam search with a beam size of 20. %Qwen3-0.6B第二次解码使用了KV-Cache存储第一次解码的隐藏状态
% For Qwen3-0.6B, the second decoding pass uses a KV-cache to reuse the hidden states from the first pass.

% All results are averaged over 3 runs on 8 NVIDIA RTX PRO 6000 GPUs. For tokenization, we extract item text embeddings by mean-pooling the last-layer hidden states of LLaMA-7B~\cite{touvron2023llama} and Qwen2.5-7B~\cite{bai2023qwen}, then discretize them with our tokenization method ($L_1{=}2,L_2{=}2$), where each level has 256 code vectors of 128 dimensions; the tokenizer is trained with AdamW~\cite{loshchilovdecoupled} (lr $=10^{-3}$, batch size $=1024$). We instantiate GenCDSR on T5~\cite{raffel2020exploring} and Qwen3-0.6B~\cite{yang2025qwen3}: T5 follows TIGER~\cite{rajput2023recommender} (AdamW lr $=5\times10^{-4}$, batch size $=256$, LoRA rank $=8$, $\alpha{=}32$), while Qwen3-0.6B uses AdamW lr $=10^{-4}$, batch size $=128$, and LoRA rank $=8$, $\alpha{=}16$. During inference, we use beam search with beam size 20; for Qwen3-0.6B, the second decoding pass reuses the first-pass hidden states via KV-cache.

\subsubsection{Evaluation Metrics}
%遵循先前的研究，本文采用 Top-$k$ 推荐评测指标评估模型性能，具体包含命中率（Hit Rate，HR@$k$）与归一化折损累计增益（Normalized Discounted Cumulative Gain，NDCG@$k$）。所有实验中，$k$ 统一取值为 10，对应指标记为 HR@10 与 NDCG@10。

% Following previous works~\cite{liu2025bridge,cao2022contrastive,wang2025nezha}, 
% we adopt Top-$k$ recommendation evaluation metrics to assess model performance, 
% including Hit Ratio (H@$k$) and Normalized Discounted Cumulative Gain (N@$k$) with $k =$ 5 and 10. 
% %In all experiments, we uniformly set $k=5, 10$, denoted as HR@5, HR@10, NDCG@5 and NDCG@10 respectively. 
% For efficiency evaluation, we additionally report the average generation latency (LT) in milliseconds (ms).

Following previous works~\cite{liu2025bridge,cao2022contrastive,wang2025nezha}, 
we adopt standard Top-$k$ recommendation metrics to evaluate model performance, 
including Hit Ratio (H@$k$) and Normalized Discounted Cumulative Gain (N@$k$) at $k\in\{5,10\}$. 
For efficiency evaluation, we further report the average generation latency (LT) measured in milliseconds (ms).

\begin{table*}[ht]
\centering
\setlength\abovecaptionskip{0\baselineskip}
\setlength\belowcaptionskip{0\baselineskip}
\caption{Overall performance comparison on all datasets. H@K and N@K denote Hit Ratio and NDCG at $k$, respectively. The best results are highlighted in bold, while the best baseline results are underlined. "\textbf{{\large *}}'' indicates statistically significant improvements over the strongest baseline according to a two-sided t-test ($p < 0.05$).}
\label{tab:overall performance}
\setlength{\aboverulesep}{0pt}
\setlength{\belowrulesep}{0pt}
\begin{tabular}{c|c|cccc|cccc|c}
\toprule
\multirow{3}{*}{\textbf{Dataset}} 
& \multirow{3}{*}{\textbf{Metric}} 
& \multicolumn{4}{c|}{\textbf{SDSR}} 
& \multicolumn{4}{c|}{\textbf{CDSR}} 
& \multirow{3}{*}{\textbf{Ours}} \\
\cline{3-10}

& 
& \multicolumn{3}{c}{\textbf{Discriminative}} 
& \multicolumn{1}{c|}{\textbf{Generative}} 
& \multicolumn{3}{c}{\textbf{Discriminative}} 
& \multicolumn{1}{c|}{\textbf{Generative}} 
&  \\
\cline{3-10}

&
& GRU4Rec & BERT4Rec & SASRec & TIGER 
& C2DSR & TriCDR & LLM4CDSR 
& GenCDR  \\
\midrule

 \multirow{4}{*}{Clothing}
&H@5  &0.3024 &0.3318 &0.3369 &0.5809& 0.5547&0.5649 &0.5586  &{\ul 0.5848}&\bm{$0.5869^{\star}$}
 \\
& H@10 & 0.3378&0.3696 &0.3752& 0.5961&0.5692& 0.5778&0.5818  & {\ul 0.6076}&\bm{$0.6097^{\star}$}
 \\
& N@5  &  0.2748&0.2994 & 0.3079 &0.5220&0.5230 &0.5324 &0.5346  & {\ul 0.5361} & \bm{$0.5373^{\star}$}
\\
\textbf{}& N@10 & 0.2816& 0.3017 & 0.3088&0.5269& 0.5277& 0.5386&0.5428  &{\ul 0.5420}& \bm{$0.5447^{\star}$}\\

% \midrule[0.1pt]
% \cline{2-11}
\cdashline{1-11}

 \multirow{4}{*}{Sports}
 &H@5  &  0.1797& 0.1868&0.1939 &0.4577& 0.4528& 0.4596&0.4668  & {\ul 0.4979}&\bm{$0.5015^{\star}$} \\
&H@10& 0.1809& 0.1876&0.1948 &0.4667&0.4885& 0.4952 &0.5038  &{\ul 0.5182}&\bm{$0.5207^{\star}$} \\
&  N@5  & 0.1134& 0.1189&0.1236&0.4484&0.3586 & 0.3664 &0.3749  &{\ul 0.4653}&\bm{$0.4671^{\star}$} \\
 & N@10 &0.1138 &0.1192 &0.1241 &0.4513&0.3668 &0.3745&0.3826  &{\ul 0.4718} &\bm{$0.4735^{\star}$} \\
\midrule

 \multirow{4}{*}{Electronics}
 &H@5  & 0.0116&0.0133 &0.0139 &0.0484&0.0448 &0.0439 & 0.0452 &{\ul 0.0531}& \bm{$0.0543^{\star}$} \\
&  H@10 &0.0154 &0.0179 &0.0186& 0.0596& 0.0525&0.0535 & 0.0567&{\ul 0.0648}&\bm{$0.0677^{\star}$} \\
 & N@5  &0.0186 & 0.0204&0.0227& 0.0401& 0.0379&0.0376 & 0.0347 &{\ul 0.0444}&\bm{$0.0447^{\star}$} \\
&  N@10 &0.0213 &0.0235& 0.0261&0.0437& 0.0404&0.0407 & 0.0384 &{\ul 0.0482}&\bm{$0.0491^{\star}$}   \\
% \cline{2-11}
\cdashline{1-11}
% \midrule

 \multirow{4}{*}{Phone}
& H@5  &0.0331 & 0.0361&0.0376 &0.0596& 0.0490&0.0509 & 0.0560 &{\ul 0.0780}&\bm{$0.0809^{\star}$}  \\
 & H@10 &0.0441 &0.0475&0.0489 &0.0786&0.0643 &0.0653& 0.0731 &{\ul 0.1045}& \bm{$0.1061^{\star}$} \\
&  N@5  &0.0256 &0.0284 & 0.0312&0.0476& 0.0408& 0.0405& 0.0403 &{\ul 0.0616}&\bm{$0.0622^{\star}$}  \\
 & N@10 &0.0291 & 0.0324&0.0356 &0.0537& 0.0457&0.0451 & 0.0457 &{\ul 0.0691}&\bm{$0.0702^{\star}$}  \\
\midrule

 \multirow{4}{*}{Book}
& H@5  &0.0124 &0.0152 &0.0178 &0.1290& 0.0201&0.0315&0.0516 &{\ul 0.1336}&\bm{$0.1350^{\star}$} \\
 & H@10&0.0198 & 0.0198&0.0208 &0.1390& 0.0372&0.0344 &0.0745 &{\ul 0.1452}& \bm{$0.1469^{\star}$} \\
&  N@5  &0.0062 &0.0081&0.0096&0.1189&0.0098&0.0205 & 0.0372 &{\ul 0.1258} &\bm{$0.1270^{\star}$}\\
 & N@10 &0.0074 &0.0092 &0.0110 &0.1222&0.0148 &0.0237 & 0.0444 &{\ul 0.1296}& \bm{$0.1308^{\star}$} \\
% \cline{2-11}
 \cdashline{1-11}
% \midrule

 \multirow{4}{*}{Movie}
& H@5  &0.0636&0.0698 & 0.0749&0.1605& 0.1472&0.1566 &  0.1621 &{\ul 0.1726}&\bm{$0.1875^{\star}$}  \\
& H@10 & 0.1070&0.1128 &0.1189&0.2473&0.2054 &0.2139& 0.2216 &{\ul 0.2622}&\bm{$0.2671^{\star}$}   \\
&  N@5 &0.0366 &0.0408&0.0449 &0.1031&0.1181 & {\ul 0.1295}
 & 0.1176 &0.1120& \bm{$0.1189^{\star}$}
 \\
 & N@10 &0.0506 &0.0548&0.0596 &0.1311&0.1284&0.1360 & 0.1344 &{\ul 0.1408}&\bm{$0.1445^{\star}$}  \\
\bottomrule
\end{tabular}
\end{table*}

\subsection{Overall Performance (RQ1)}
Table~\ref{tab:overall performance} presents the overall performance comparison between GenCDSR and all baseline methods across three cross-domain datasets. Based on the experimental results, several key observations can be derived.
First, GenCDSR achieves the best performance across most datasets and evaluation metrics. This indicates that by coupling generative modeling with cross-domain knowledge transfer via our cross-domain hybrid tokenization and serial--parallel decoding mechanism, it effectively enhances cross-domain recommendation capability.
Second, cross-domain signals effectively enhance single-domain representations, thereby alleviating data sparsity. In general, CDSR models outperform SDSR baselines, suggesting that information from related domains provides complementary supervision and helps the model learn more robust user preferences from limited target-domain interactions.
Third, generative models are more suitable for cross-domain scenarios. Compared with discriminative approaches, they are better at modeling sequential dependencies and transferable semantic patterns across domains. This indicates that generative frameworks are better aligned with the goal of cross-domain recommendation, namely, capturing both shared knowledge and domain-specific characteristics.

\subsection{Efficiency Analysis (RQ2)} \label{sec:efficiency}
%effency：概括性结论：分别在t5和相对beam search三个任务和qwenbackbone三个任务相比beam search（LT）；相对于mtp和nezha效果提升，insummry兼顾相比于beam search来比；写数字

%\paragraph{Decoding}
%在组件分析实验中，我们进一步对比不同的解码策略，以评估所提出混合解码设计的有效性。具体而言，我们考虑以下解码变体：
%In the component analysis, we further compare different decoding strategies to evaluate the effectiveness of our proposed serial-parallel decoding design. We evaluate the following baseline decoding variants:
We compare our proposed serial-parallel decoding strategy with the following SOTA decoding strategies on T5 and Qwen3-0.6B.
\begin{itemize}[leftmargin=*]
\item \textbf{Beam Search}~\cite{freitag2017beam} is a standard token-by-token autoregressive decoding strategy which ensures grounded item.
\item \textbf{Multi-token Prediction (MTP)}~\cite{gloeckle2024better} is a multi-token prediction strategy that predicts multiple tokens in parallel.
\item \textbf{NEZHA}~\cite{wang2025nezha} uses a self-drafting head with hash-set verification for fast SID decoding to improve efficiency.
\end{itemize}
\begin{table*}[ht]
\centering
\setlength\abovecaptionskip{0\baselineskip}
\setlength\belowcaptionskip{0\baselineskip}
\caption{Efficiency analysis of the proposed cross-domain serial-parallel decoding. H@K, N@K, and LT denote Hit Ratio, NDCG at $k$, and average generation latency (in milliseconds), respectively. The best results are highlighted in bold. } %and second-best results are underlined. $\star$ denotes statistically significant improvements according to a two-sided t-test ($p < 0.05$).
\label{tab:Component analysis}
\setlength{\tabcolsep}{4.5pt}
\renewcommand{\arraystretch}{1}
\begin{tabular}{c|c|cccc|cccc}
\hline
\multirow{2}{*}{\textbf{Dataset}}  & \multirow{2}{*}{\textbf{Metric}} 
& \multicolumn{4}{c|}{\textbf{T5}} 
& \multicolumn{4}{c}{\textbf{Qwen3-0.6B}} \\
\cline{3-10}
 & 
 & MTP & NEZHA & Beam Search & Ours 
 & MTP & NEZHA & Beam Search & Ours  \\
\hline
\multirow{5}{*}{Clothing}
 & H@5  &0.5674 &0.5719 &\bm{$0.6013$} &{ 0.5869}  &0.5937 &0.6009 &\bm{$0.6153$} &{ 0.6046} \\
 & H@10 &0.5833 &0.5811 &\bm{$0.6227$} &{ 0.6097}  &0.6052 &0.6082 &\bm{$0.6348$} &{0.6209} \\
 & N@5  &0.5204 &0.5236 &\bm{$0.5391$} &{0.5373}  &0.5703 &0.5799 &\bm{$0.5875$} &{ 0.5807} \\
 & N@10 &0.5255 &0.5266 &\bm{$0.5461$} &{0.5447}  &0.5740 &0.5823 &\bm{$0.5939$} &{0.5860} \\
 & LT$\downarrow$ &0.1827 &0.3110 &4.6008 &0.4835 &0.5662 &1.3190 &47.2983 &6.4449 \\
\cdashline{1-10}

\multirow{5}{*}{Sports}
 & H@5  &0.4817 &0.4793 &{ 0.4847} &\bm{$0.5015$} &0.5068 &0.5062 &{0.5110} &\bm{$0.5159$} \\
 & H@10 &0.4967 &0.4907 &{0.5008} &\bm{$0.5207$} &0.5096 &0.5092 &{0.5236} &\bm{$0.5297$} \\
 & N@5  &0.4616 &0.4578 &{0.4637} &\bm{$0.4671$} &0.4906 &0.4932 &{0.4980} &\bm{$0.5006$} \\
 & N@10 &0.4634 &0.4605 &{0.4690} &\bm{$0.4735$} &0.4965 &0.4942 &{0.5020} &\bm{$0.5049$} \\
 & LT$\downarrow$ &0.3589 &0.5561 &4.6022 &0.6754 &1.3290 &1.3366 &31.2637 &6.5448 \\
\hline

\multirow{5}{*}{Electronics}
 & H@5  &0.0381 &0.0294 &{0.0529} &\bm{$0.0543$} &{ 0.0534} &0.0517 &0.0522 &\bm{$0.0574$} \\
 & H@10 &0.0463 &0.0316 &{0.0675} &\bm{$0.0677$} &0.0616 &0.0568 &{0.0632} &\bm{$0.0683$} \\
 & N@5  &0.0315 &0.0258 &{0.0435} &\bm{$0.0447$} &{0.0474} &0.0460 &0.0440 &\bm{$0.0492$} \\
 & N@10 &0.0342 &0.0265 &{0.0482} &\bm{$0.0491$} &{ 0.0489} &0.0476 &0.0475 &\bm{$0.0527$} \\
 & LT$\downarrow$ &0.1481 &0.3061 &4.5600 &0.4425 &1.4098 &1.2075 &40.7079 &6.7692 \\
\cdashline{1-10}

\multirow{5}{*}{Phone}
 & H@5  &0.0557 &0.0559 &{0.0769} &\bm{$0.0809$} &0.0740 &{0.0794} &0.0772 &\bm{$0.0834$} \\
 & H@10 &0.0735 &0.0690 &\bm{$0.1066$} &{0.1061} &0.0941 &0.0958 &\bm{$0.1057$} &{0.1021} \\
 & N@5  &0.0468 &0.0479 &{0.0611} &\bm{$0.0622$} &0.0586 &{0.0642} &0.0605 &\bm{$0.0678$} \\
 & N@10 &0.0526 &0.0522 &\bm{$0.0706$} &{0.0702} &0.0651 &0.0695 &{0.0697} &\bm{$0.0738$} \\
 & LT$\downarrow$ &0.1983 &0.3999 &4.4747 &0.5286 &1.1914 &1.2736 &27.3186 &6.8506 \\
\hline

\multirow{5}{*}{Book}
 & H@5  &0.1032 &0.0556 &{0.1330} &\textbf{0.1350} &0.1181 &0.1132 &\bm{$0.1588$} &{0.1400} \\
 & H@10 &0.1062 &0.0585 &{ 0.1420} &\bm{$0.1469$} &0.1241 &0.1132 &\bm{$0.1658$} &{ 0.1519} \\
 & N@5  &0.1025 &0.0483 &\bm{$0.1274$} &{ 0.1270} &0.1109 &0.1095 &\bm{$0.1417$} &{0.1310} \\
 & N@10 &0.1034 &0.0494 &{ 0.1303} &\bm{$0.1308$} &0.1128 &0.1095 &\bm{$0.1439$} &{ 0.1349} \\
 & LT$\downarrow$ &0.3953 &0.6436 &5.0300 &0.8085 &2.2076 &1.6720 &50.6445 &6.9210 \\
\cdashline{1-10}

\multirow{5}{*}{Movie}
 & H@5  &0.1210 &0.1147 &\bm{$0.1851$} &{0.1793} &0.1885 &0.2126 &{0.2685} &\bm{$0.2821$} \\
 & H@10 &0.1600 &0.1234 &\bm{$0.2675$} &{ 0.2671} &0.2131 &0.2454 &{0.3336} &\bm{$0.3481$} \\
 & N@5  &0.0831 &0.0700 &\bm{$0.1183$} &{0.1154} &0.1597 &0.1710 &{0.2040} &\bm{$0.2190$} \\
 & N@10 &0.0958 &0.0771 &\bm{$0.1450$} &{ 0.1445} &0.1676 &0.1816 &{0.2251} &\bm{$0.2407$} \\
 & LT$\downarrow$ &0.2565 &0.4312 &5.0010 &0.6045 &1.9899 &1.4696 &50.2618 &7.0854 \\
\hline
\end{tabular}
\end{table*}

Table~\ref{tab:Component analysis} summarizes the results and yields the following observations. 
Overall, our method achieves the best accuracy-latency trade-off: it consistently outperforms MTP and NEZHA on most datasets and metrics, with an average relative accuracy gain of 21.9\% over MTP and 37.9\% over NEZHA, averaged across all metrics and both backbones. At the same time, it remains competitive with Beam Search in recommendation quality, indicating that the proposed hybrid serial-parallel decoding largely preserves generation fidelity while substantially lowering decoding cost in practice. 
Moreover, although Beam Search attains strong accuracy, it incurs much higher latency; in contrast, our decoding reduces LT by 85.1\% on average compared with Beam Search (87.5\% on T5 and 82.7\% on Qwen3-0.6B), making it notably more practical for real-time deployment. 
In summary, the proposed hybrid serial-parallel decoding aligns well with the two-stage SID structure and offers a more favorable accuracy-latency balance by significantly reducing inference overhead without sacrificing generation quality overall.

To further interpret the efficiency results, we analyze per-position prediction accuracy (Hit@1) on the Electronics and Phone domains. On Electronics, MTP and NEZHA achieve average per-head accuracies of 0.0538 and 0.0499, respectively, compared with 0.0598 for Ours; on Phone, their averages (0.0964 and 0.0955) are also lower than Ours (0.0969). This gap mainly stems from the fully parallel decoding adopted by NEZHA and MTP, which predicts each position independently and cannot effectively leverage preceding token semantics, whereas Ours preserves partial serial dependencies while enabling parallel prediction, yielding a better balance between per-position accuracy and inference efficiency.

\subsection{Ablation Study  (RQ3)}
\begin{figure*}[t] 
\centering 
\setlength\abovecaptionskip{0.2\baselineskip}
\setlength\belowcaptionskip{0.2\baselineskip}
\begin{tabular}{cc}
    % 左侧区域：Ablation Study
    \begin{minipage}[t]{0.60\textwidth} 
        \centering 
        \includegraphics[width=0.85\linewidth]{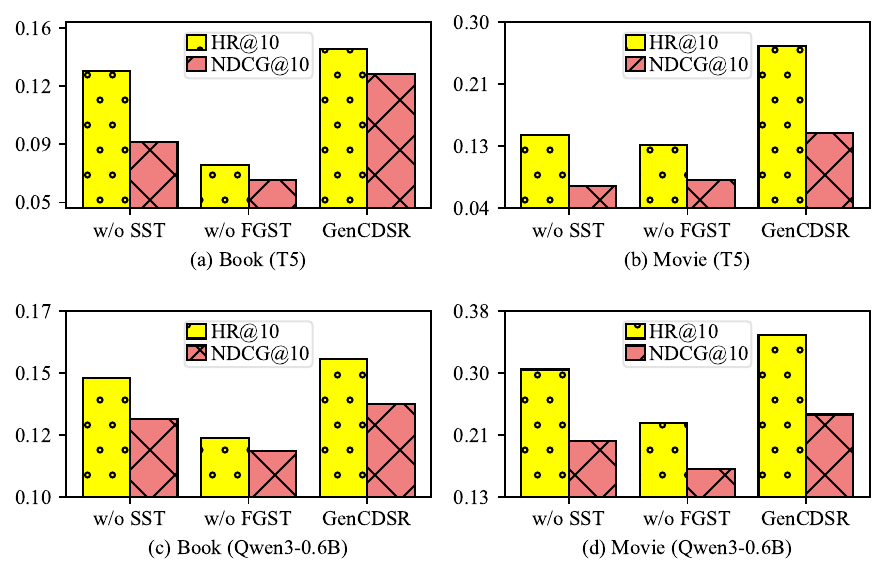} 
        % \vspace{1mm} 
        \label{fig:ablation}
    \end{minipage} 
    &
    % 右侧区域：Feature Fidelity
    \begin{minipage}[t]{0.38\textwidth} 
        \centering 
        \raisebox{0.5cm}{\includegraphics[width=0.85\linewidth]{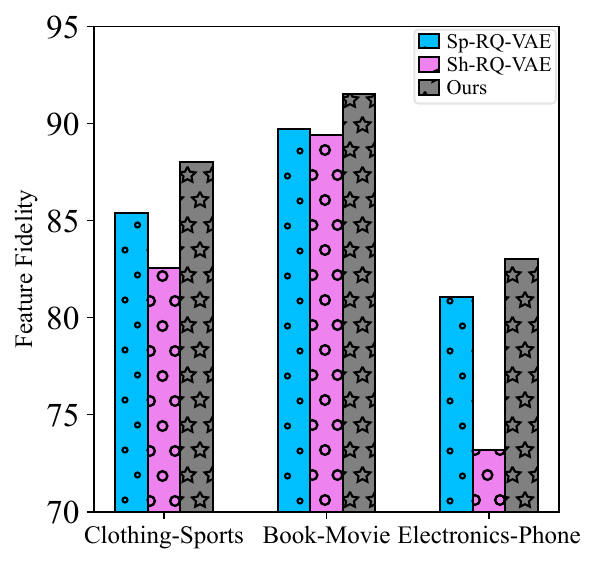}}
        
        \vspace{-6mm}
        \small (e) Feature fidelity comparison (higher is better). 
    \end{minipage}
    \vspace{-10pt}
\end{tabular}

% \vspace{-8mm}
\caption{Ablation study and feature fidelity comparison (higher is better for all subfigures). (a-d) Ablation results on the Book-Movie dataset under T5 and Qwen3-0.6B; SST and FGST denote Shared-Specific Tokenization and Fine-Grained Specific Tokenization, respectively. (e) Feature fidelity comparison on three cross-domain datasets.}
\label{fig:ablation_fidelity}
\vspace{-10pt}
\end{figure*}

Figure~\ref{fig:ablation_fidelity} (a-d) reports the ablation results of the proposed cross-domain hybrid tokenization on the Book-Movie dataset. The variants w/o Shared-Specific Tokenization and w/o Fine-Grained Specific Tokenization remove Stage~1 and Stage~2 tokenization modules, respectively.
Across both T5 and Qwen3-0.6B backbones, the full model (GenCDSR) achieves the best performance on H@10 and N@10, with most gains being statistically significant. Removing Stage~1 causes larger performance drops, highlighting its role in learning transferable shared semantics for cross-domain modeling, while removing Stage~2 mainly hurts domain-level precision, indicating the necessity of fine-grained domain-specific refinement. Overall, the two stages are complementary, and their joint design is crucial for robust gains.

% \begin{figure}[t]

%     \centering
%     %\includegraphics[width=3in]{fig5}
%     % \subfloat[Clothing]{
%     % 		\includegraphics[scale=0.275]{Figure/ablation/clothing_t5_ablation.pdf}\label{fig:ratio_clothing}}
%     % \subfloat[Sports]{
%     % 		\includegraphics[scale=0.275]{Figure/ablation/sports_t5_ablation.pdf}\label{fig:ratio_sports}}
%     % \subfloat[Clothing]{
%     % 		\includegraphics[scale=0.275]{Figure/ablation/clothing_qwen3-06b_ablation.pdf}\label{fig:ratio_clothing}}
%     % \subfloat[Sports]{
%     % 		\includegraphics[scale=0.275]{Figure/ablation/sports_qwen3-06b_ablation.pdf}\label{fig:ratio_sports}}
%     % \\
%     \subfloat[Book]{
%     		\includegraphics[scale=0.275]{Figure/ablation/books_t5_ablation.pdf}\label{fig:ratio_book}}
%     \subfloat[Movie]{
%     		\includegraphics[scale=0.275]{Figure/ablation/movies_t5_ablation.pdf}\label{fig:ratio_movie}}\\
%     \subfloat[Book]{
%     		\includegraphics[scale=0.275]{Figure/ablation/books_qwen3-06b_ablation.pdf}\label{fig:ratio_clothing}}
%     \subfloat[Movie]{
%     		\includegraphics[scale=0.275]{Figure/ablation/movies_qwen3-06b_ablation.pdf}\label{fig:ratio_sports}}
    
%     \caption{Ablation study of the proposed cross-domain hybrid tokenization on two datasets (higher is better).}
%     \label{fig:ablation}
% \end{figure}

\subsection{Feature Fidelity Analysis (RQ4)} \label{sec:ffa}
%如图所示，我们比较了三种 tokenization 方案在三个跨域数据集上的特征保真度（Feature Fidelity），包括 RQ-VAE（跨域共享同一个 RQ-VAE）、Domain-RQ-VAE（每个域独立训练一个 RQ-VAE）以及 Ours。整体结果表明，Ours 在所有数据集上均取得最高的保真度，说明所生成的语义 ID 能更充分地保留原始语义特征信息。相比之下，RQ-VAE 的跨域共享容易在域间异质性较强时引入信息损失，而 Domain-RQ-VAE 虽缓解了域内差异带来的干扰，但缺乏跨域共享结构的约束，整体仍不及本文方法。总体而言，该结果验证了本文的 share–specific tokenization 在离散化过程中实现了更好的信息保留与表示稳定性，为后续生成式推荐建模提供了更可靠的表征基础。
To illustrate how the cross-domain hybrid tokenization contributes to improving recommendation performance, we evaluate the feature fidelity~\cite{fu2025forge} of different tokenization schemes on three cross-domain datasets, which is defined as:
\begin{equation}
\setlength{\abovedisplayskip}{0pt}
\setlength{\belowdisplayskip}{0pt}
\label{eq:feature_fidelity_joint}
\text{Feature Fidelity}
=
\max\!\left(
0,\,
1-\frac{\left\|[\mathbf{x}_A;\mathbf{x}_B]-[\hat{\mathbf{x}}_A;\hat{\mathbf{x}}_B]\right\|_2}
{\left\|[\mathbf{x}_A;\mathbf{x}_B]\right\|_2}
\right)\times 100\%.
\end{equation}
where $\mathbf{x}_A,\mathbf{x}_B$ are the original semantic embeddings for domain $A$ and $B$, $\hat{\mathbf{x}}_A,\hat{\mathbf{x}}_B$ are the reconstructed embeddings, $[\cdot;\cdot]$ denotes concatenation operation, and $\|\cdot\|_2$ denotes the $\ell_2$ norm.

We compare our method with Sh-RQ-VAE (a single shared RQ-VAE across domains) and Sp-RQ-VAE (one RQ-VAE per domain). As shown in Figure~\ref{fig:ablation_fidelity}(e), our method consistently achieves the highest feature fidelity on all datasets, indicating more faithful preservation of semantic features in the learned SIDs. In contrast, Sh-RQ-VAE is more prone to information loss when inter-domain heterogeneity is large, while Sp-RQ-VAE reduces intra-domain interference but lacks an explicit shared structure across domains. Overall, these results further confirm that our tokenization preserves information more effectively and produces more stable discretized representations for downstream generative recommendation, while better supporting subsequent cross-domain knowledge transfer and sequence modeling across heterogeneous domains.

\subsection{Hyper-Parameter Analysis (RQ5)}
\begin{figure}[t]
\setlength\abovecaptionskip{0.2\baselineskip}
\setlength\belowcaptionskip{0.2\baselineskip}
    \centering
    \includegraphics[width=0.85\linewidth]{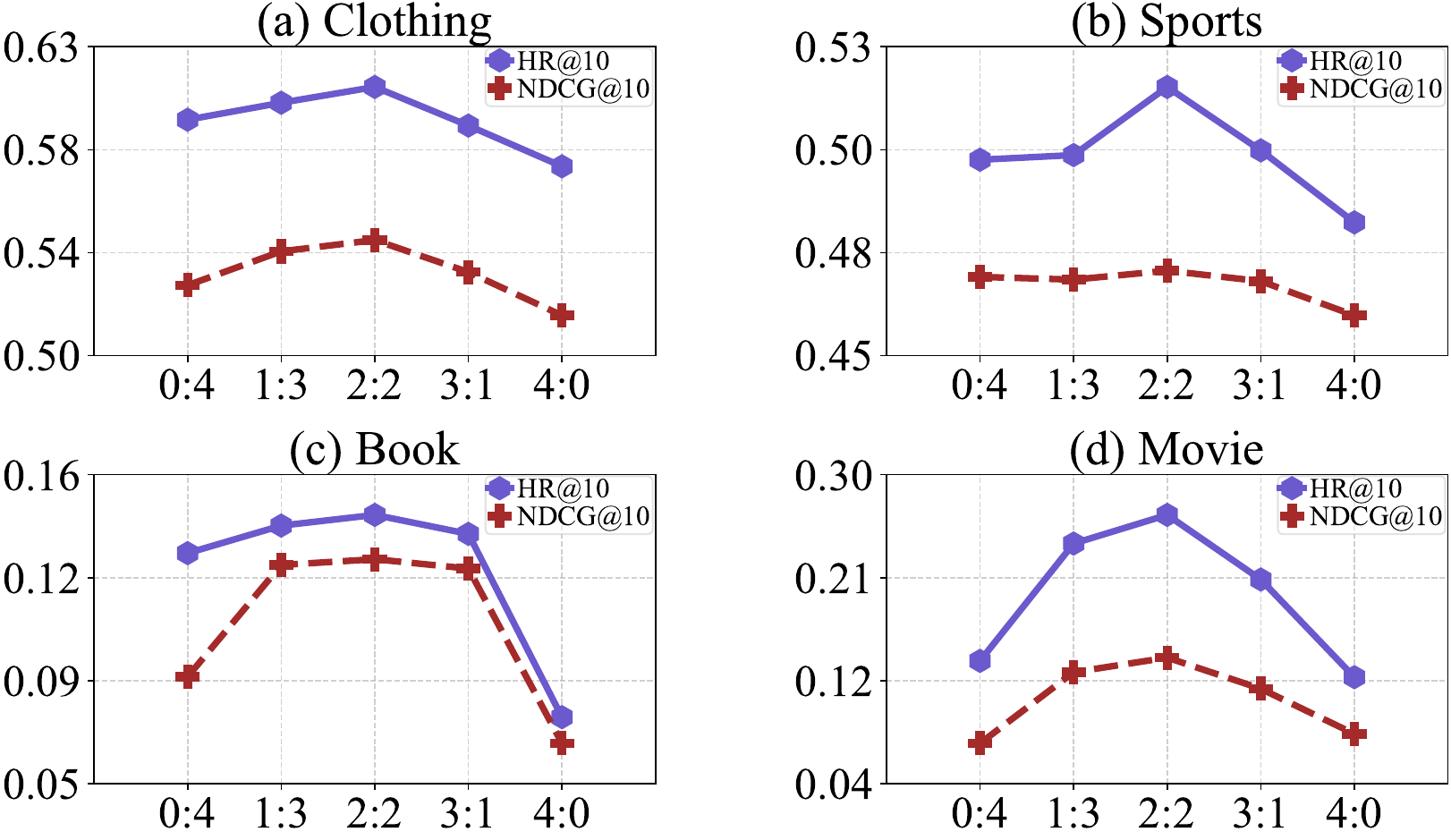}
    \caption{Impact of the codebook allocation between shared-specific tokenization and fine-grained specific tokenization (total = 4) on HR@10 and NDCG@10 across two datasets.}
     \label{fig:ratio}
     \vspace{-20pt}
\end{figure}

We set the code level to $L=4$ and vary the shared-specific tokenization vs.\ fine-grained specific tokenization allocation in $\{0\!:\!4,\,1\!:\!3,\,2\!:\!2,\,3\!:\!1,\,4\!:\!0\}$ to evaluate cross-domain hybrid tokenization. As shown in Figure~\ref{fig:ratio}, $2\!:\!2$ achieves the best performance, suggesting that balanced capacity is crucial under our two-stage design across datasets and metrics. Too little shared capacity (e.g., $0\!:\!4$ or $1\!:\!3$) weakens Stage~1 cross-domain anchoring and alignment, while too much shared capacity (e.g., $3\!:\!1$ or $4\!:\!0$) suppresses Stage~2 fine-grained refinement and may increase semantic ambiguity. Overall, $2\!:\!2$ offers the best trade-off between shared structure and domain-specific refinement in practice under all settings.
% \begin{figure}[t]

%     \centering
%     %\includegraphics[width=3in]{fig5}
%     \subfloat[Clothing]{
%     		\includegraphics[scale=0.25]{Figure/ratio_line/ratio_clothing.pdf}\label{fig:ratio_clothing}}
%     \subfloat[Sports]{
%     		\includegraphics[scale=0.25]{Figure/ratio_line/ratio_sports.pdf}\label{fig:ratio_sports}}
%     \\
%     \subfloat[Book]{
%     		\includegraphics[scale=0.25]{Figure/ratio_line/ratio_book.pdf}\label{fig:ratio_book}}
%     \subfloat[Movie]{
%     		\includegraphics[scale=0.25]{Figure/ratio_line/ratio_movie.pdf}\label{fig:ratio_movie}}
    
%    \caption{Impact of the codebook allocation between Share--Specific Tokenization and Fine-Grained Specific Tokenization (total = 4) on HR@10 and NDCG@10 across four datasets.}
%     \label{fig:ratio}
% \end{figure}

\section{Related Work}

\subsection{Cross-Domain Sequential Recommendation }
CDSR jointly models users’ interaction sequences across multiple domains to capture cross-domain interest transitions and improve next-item prediction~\cite{chen2024survey}. Compared with single-domain sequential recommendation~\cite{liu2024multimodal,wang2023single,liu2023multi} and cross-domain recommendation~\cite{li2023hamur,li2022gromov,jia2024d3,gao2023autotransfer}, it faces substantial domain heterogeneity, imbalanced interactions, and complex cross-domain transitions~\cite{wang2019sequential}. Most methods follow collaborative filtering to transfer knowledge by modeling user/item relations and cross-domain dependencies~\cite{zhu2022personalized,li2020ddtcdr}, including graph neural network-based approaches~\cite{wu2022graph} such as C2DSR~\cite{cao2022contrastive} and contrastive variants for stronger representation consistency~\cite{wang2023unbiased,xu2025multi}. Recent work like TriCDR~\cite{ma2024triple} also models mixed behavior sequences and leverages LLMs like LLM4CDSR~\cite{liu2025bridge} and LLM-EDT~\cite{liu2025llm}; however, these methods still largely rely on collaborative signals and make limited use of item-level semantics and explicit cross-domain semantic relationships. %~\cite{achiam2023gpt,touvron2023llama,bai2023qwen} (

\subsection{Generative Recommendation}
Generative methods have recently gained prevalence across various fields~\cite{zhang2026tearag,han2026data,10.1145/3627673.3679687,chen2025mghft}. Generative recommendation (GR) reformulates recommendation as autoregressive sequence generation, providing a unified pipeline~\cite{ji2024genrec,li2023gpt4rec,li2025ctr,xiong2024dq,
li2024uncertaintyrag} that typically consists of item tokenization and recommendation generation. For item tokenization, quantization techniques are commonly employed to map items into discrete semantic identifiers (SIDs). Residual Quantization, such as RQ-VAE~\cite{rajput2023recommender,wang2024learnable}, and Product Quantization (PQ)~\cite{luo2025qarm,zhang2024learning} are widely adopted to produce structured codewords. For example, TIGER~\cite{rajput2023recommender} utilizes RQ-VAE to construct hierarchical SIDs, while LC-Rec~\cite{zheng2024adapting} employs learnable codebooks to enhance semantic alignment. Building on these advances, GenCDR~\cite{hu2026ids} and MTCDR~\cite{jin2025generative} further extend such tokenization paradigms to multi-domain settings. Regarding recommendation generation, most methods rely on next-token prediction with LLM backbones, while increasing attention has been devoted to decoding efficiency. For example, NEZHA~\cite{wang2025nezha} introduces a multi-token prediction strategy for SID decoding to alleviate the high inference latency of sequential generation. However, existing approaches still insufficiently capture cross-domain collaborative signals during tokenization and suffer from the accuracy--efficiency trade-off during decoding.

% \wyh{Some works targets at tokenization. }
\section{Conclusion}

In this paper, we propose GenCDSR, an effective and efficient generative framework for cross-domain sequential recommendation. It employs cross-domain hybrid tokenization with shared-specific and fine-grained specific tokenization to capture cross-domain commonalities and distinctions. It further introduces serial--parallel decoding to balance recommendation accuracy and inference efficiency. Experiments on real-world cross-domain datasets demonstrate that GenCDSR outperforms SOTA baselines, improving accuracy by 1.5\% while reducing inference latency by 85.1\%.

\section{Acknowledgments}
This research was partially supported by National Natural Science Foundation of China (No.62502404), Hong Kong Research Grants Council (Research Impact Fund No.R1015-23, Collaborative Research Fund No.C1043-24GF, RGC Research Fellow Scheme No.RFS2627-1S03, General Research Fund No. 11218325, No. 11212926), Institute of Digital Medicine of City University of Hong Kong (No.9229503), and Bytedance.
% \nocite{*}
\bibliographystyle{ACM-Reference-Format}
\bibliography{8.reference} 

\end{document}